\documentclass[british,onecolumn]{IEEEtran}
\usepackage[T1]{fontenc}
\usepackage[utf8]{inputenc}
\usepackage{float}
\usepackage{textcomp}
\usepackage{url}
\usepackage{graphicx}

\makeatletter

\providecommand{\tabularnewline}{\\}

\usepackage{graphicx}

\usepackage{booktabs}
\usepackage[style=ieee,
    citestyle=numeric-comp,
    hyperref=false,%true,
    url=false, %true,
    backref=false, %true,
    natbib=true,
    backend=biber]{biblatex}
\providecommand{\etal}[0]{\textit{et al.} }

\makeatother

\usepackage{babel}
\begin{document}

\title{Enabling Fingerprint Presentation Attacks: Fake Fingerprint Fabrication
Techniques and Recognition Performance}

\author{\IEEEauthorblockN{Christof Kauba, Luca Debiasi and Andreas Uhl\\} 
\IEEEauthorblockA{Department of Computer Sciences\\ University of Salzburg\\ Jakob-Haringer-Str. 2\\ 5020 Salzburg\\ AUSTRIA\\ Email:  \{ckauba, ldebiasi, uhl\}@cs.sbg.ac.at} }
\maketitle
\begin{abstract}
Fake fingerprint representation pose a severe threat for fingerprint
based authentication systems. Despite advances in presentation attack
detection technologies, which are often integrated directly into the
fingerprint scanner devices, many fingerprint scanners are still susceptible
to presentation attacks using physical fake fingerprint representation.
In this work we evaluate five different commercial-off-the-shelf fingerprint
scanners based on different sensing technologies, including optical,
optical multispectral, passive capacitive, active capacitive and thermal
regarding their susceptibility to presentation attacks using fake
fingerprint representations. Several different materials to create
the fake representation are tested and evaluated, including wax, cast,
latex, silicone, different types of glue, window colours, modelling
clay, etc. The quantitative evaluation includes assessing the fingerprint
quality of the samples captured from the fake representations as well
as comparison experiments where the achieved matching scores of the
fake representations against the corresponding real fingerprints indicate
the effectiveness of the fake representations. Our results confirmed
that all except one of the tested devices are susceptible to at least
one type/material of fake fingerprint representations.
\end{abstract}


\section{Introduction}

In our modern world there is an ever growing need to identify individuals,
especially for personal authentication, e.g. to unlock a smart phone,
to authorize a banking transaction or to withdraw some money from
an automated teller machine. Biometric recognition systems provide
a convenient way to perform this necessary authentication step without
having to hassle with keys, smart cards or having to remember complicated
passwords and thus, have the potential to provide some extra security
as well. According to the ISO/IEC standard TR24741:2018, biometric
recognition is defined as ``the automated recognition of individuals
based on their biological and behavioural characteristics.'' \cite{ISOIECTR24741-2018}.
The most commonly used biological traits include fingerprints, face,
iris and voice. For each of those traits, a special device to capture
samples of the particular trait is needed, commonly denoted as biometric
capturing device or formerly as biometric scanner. Most modern fingerprint
scanners are based on optical or capacitive technology and can be
built as small as a one cent coin, enabling the integration of fingerprint
recognition technology in a great variety of different devices, from
door locks over laptop computer to smartphones. Capturing a fingerprint
small is a quick and reliable process which enjoys a high users' acceptance.
All those advantages lead to the widespread utilisation of fingerprint
recognition and made fingerprints the predominant biometric trait.

Despite those advantages, fingerprint recognition technology is far
from being perfect. In contrast to passwords and tokens, a biometric
trait can neither be changed nor revoked. Thus the legitimate users
of a biometric recognition system are put in a dangerous situation,
if biometric data gets compromised or is abused. Researchers found
several different vulnerabilities of biometric recognition systems
in general and fingerprint systems in particular. These attack possibilities
or vulnerabilities can be grouped into eight different attack points
\cite{BRatha01a}: (1) presentation attack, (2) biometric signal replication,
(3) feature modification, (4) replacing features, (5) overriding the
matcher, (6) replacing templates, (7) modifying data through the channel
and (8) altering the decision. Except the first one, the presentation
attack, all remaining 7 attack possibilities refer to alterations
in digital signals or digitally stored information, while the first
one refers to a ``physical'' or sensor-level attack to the biometric
capturing device by presenting a ``spoof'' or ``fake'' representation
which is a reproduction of a genuine biometric trait. These remaining
7 attack possibilities can be prevented by employing encryption and
device authentication on the transmission channels as well as template
protection for the database. The presentation attack however, is a
direct physical attack against the sensor and one of the most important
attack possibilities \cite{BEspinoza11a,BEspinoza11b} as the attack
target (the sensor) is available, it is easy to implement without
further knowledge about the internals of the biometric recognition
system and since there exist several well documented and easy to realise
methods for fake fingerprint production which do not require any particular
prior knowledge and are not cost expensive. In the following we will
focus on this particular attack, the presentation attack in the scope
of fingerprint recognition.

As fingerprint systems are regarded as highly reliable and are often
used to secure confidential and important information, a successful
presentation attack (i.e. the biometric recognition system does not
detect the attack and the fake representation qualifies the fake as
genuine sample of the corresponding subject) poses a severe threat
to fingerprint recognition systems. By launching a successful presentation
attack, an adversary can gain unauthorized access to a system by ``impersonating''
a genuine user and then access the confidential data. Several different
methods of generating fake fingerprint representation (also denoted
as spoofing artefact) have been reported, most of them using materials
like gelatine, wax, wood glue, moldable plastics, clay, dental impression
paste and silicone. A detailed review on those materials is given
in Section \ref{sec:Literature-Review}. Due to the high risk originating
from successful presentation attacks, researchers and industry implemented
ways to detect and/or prevent presentation attacks in order to secure
their biometric systems. Those systems can either be implemented in
hardware or software and are denoted as presentation attack detection
(PAD) systems. PAD systems are out of the scope of this work.

The rest of this paper is organised as follows: Section \ref{sec:Fingerprint-Recognition}
provides a short overview on fingerprint recognition and the fingerprint
features used to identify an individual. Furthermore it summarised
the different types of fingerprint capturing devices. Section \ref{sec:Literature-Review}gives
a literature review on different fingerprint presentation attack methods
and different fake fingerprint fabrication materials. In Section
\ref{sec:Experimental-Evaluation} the fabrication of different types
of fake fingerprint representations is described and an experimental
evaluation of several commercial-off-the-shelf (COTS) fingerprint
capturing devices and using the fake fingerprint representations is
conducted. Finally, Section \ref{sec:Conclusion} concludes this paper.

\section{Fingerprint Recognition\label{sec:Fingerprint-Recognition}}

A typical biometric recognition system consists of the following modules/stages:
acquisition of the sample, pre-processing, feature extraction, comparison,
final decision \cite{BMaltoni03a}. It works in two phases. The first
one is the enrolment, where one or multiple samples of each subject
are captured, pre-processed and the extracted features are stored
as biometric templates in a database. The second one is the verification/identification,
where a new biometric sample is captured, again pre-processed and
the extracted features are compared against the templates of the corresponding
subject (verification) or against all templates in the database (identification)
to arrive at the final decision (genuine or impostor in case of verification
and a rank of candidate matches in identification). In fingerprint
recognition systems, typically three to five samples are captured
in order to derive a template that is independent of variations in
placement, deformation and rotation of the fingertip. 

\begin{figure}
\begin{centering}
\includegraphics[width=0.35\columnwidth]{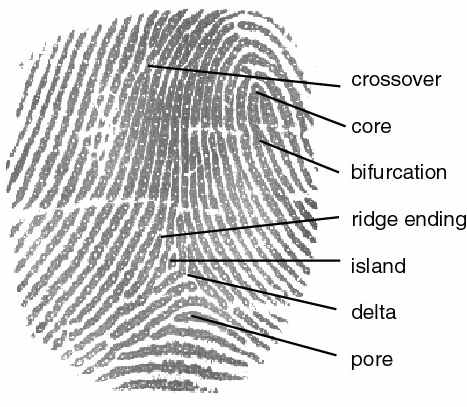}
\par\end{centering}
\caption{Typical fingerprint example. Image taken from \protect\url{http://cnx.org/content/m12574/latest/}\label{fig:Fingerprint-example}}
\end{figure}

\begin{figure}
\centering{}\includegraphics[width=1\columnwidth]{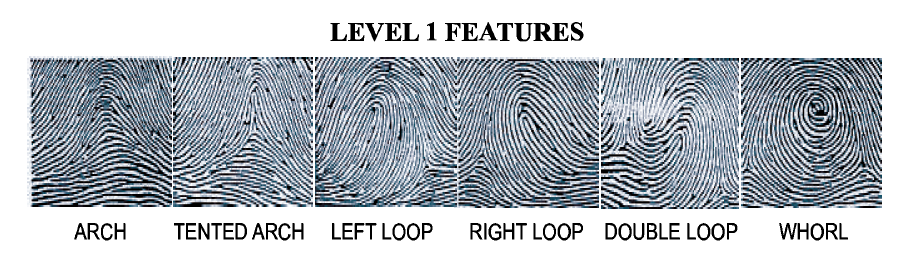}\caption{Level 1 fingerprint details - archs, loops and whorls. Image taken
from \cite{BAbhishek15a}.\label{fig:Level-1-fingerprint-details}}
\end{figure}

\begin{figure}
\centering{}\includegraphics[width=1\columnwidth]{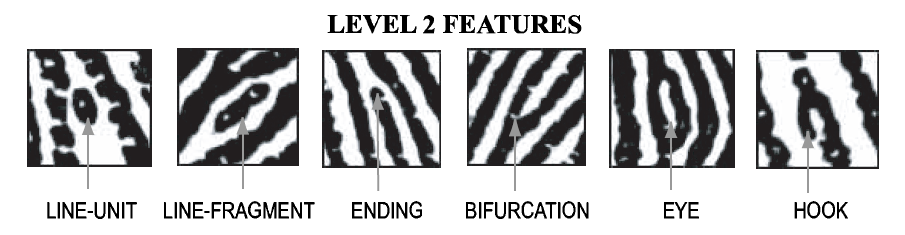}\caption{Level 2 fingerprint details - ridge endings, bifurcations, eyes, hooks,
line units and line fragments. Image taken from \cite{BAbhishek15a}.\label{fig:Level-2-fingerprint-details}}
\end{figure}

\begin{figure}
\centering{}\includegraphics[width=1\columnwidth]{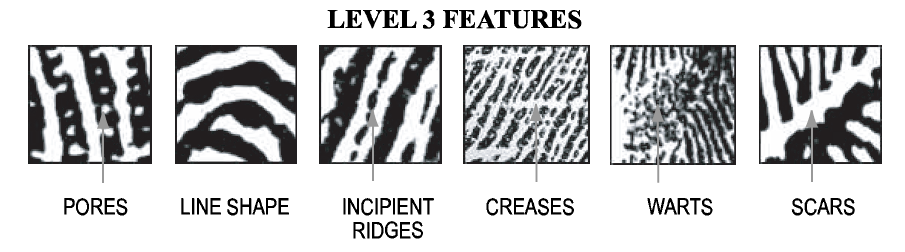}\caption{Level 3 fingerprint details - pores, line shapes, incipient ridges,
creases, wrats and scars. Image taken from \cite{BAbhishek15a}.\label{fig:Level-3-fingerprint-details}}
\end{figure}

The skin on the inside of a finger is covered with a pattern of ridges
and valleys. These ridges are not continuous but can either end, fork
or form an island. These ridge patterns are believed to be unique
for each person and relatively stable over time. According to Maltoni
et al. \cite{BMaltoni03a} a fingerprint refers to ``a flowing pattern
consisting of ridges and valleys on the fingertip of an individual''.
Figure \ref{fig:Fingerprint-example} shows a typical impression of
a fingerprint with some of the main features named and highlighted.
Most fingerprint recognition systems rely on specific characteristics
in the pattern of the ridges, which can be separated in three different
levels:
\begin{enumerate}
\item Global level (see figure \ref{fig:Level-1-fingerprint-details}):
Macro details such as the pattern or type of ridges and valleys are
detected. Ridges exhibit several regions where they resemble a distinct
shape, usually classified into deltas, loops and whorls.
\item Local level (see figure \ref{fig:Level-2-fingerprint-details}): Minutiae
points, which are the major feature to describe fingerprints, describe
different anomalies in the ridge structure, like ridge endings and
bifurcations. A minutiae point is represented by its location (x and
y coordinate), the ridge direction (angle $\theta$) and the type
of anomaly (ending or bifurcation).
\item Very fine level (see figure \ref{fig:Level-3-fingerprint-details}):
Small details such as incipient ridges and sweat pores can be detected
if the capturing devices has a sufficient resolution (1000 dpi at
least) and utilised as well. Pores can be further classified into
open or closed, depending on their position on the ridges. Currently,
those level 3 features are much harder to forge than level 1 and level
2 features.
\end{enumerate}
A typical fingerprint contains about one hundred minutiae points.
The area that is captured by usual fingerprint sensors contains about
30-50 minutiae points. For a positive identification that stands in
European courts, at least 12 minutiae have to be unambiguously identified
in a fingerprint, while most commercial fingerprint recognition systems
are able to perform a successful positive match with a minimum of
8 minutiae.

\subsection{Fingerprint Sensing Technologies}

\begin{figure}
\begin{centering}
\includegraphics[width=0.75\columnwidth]{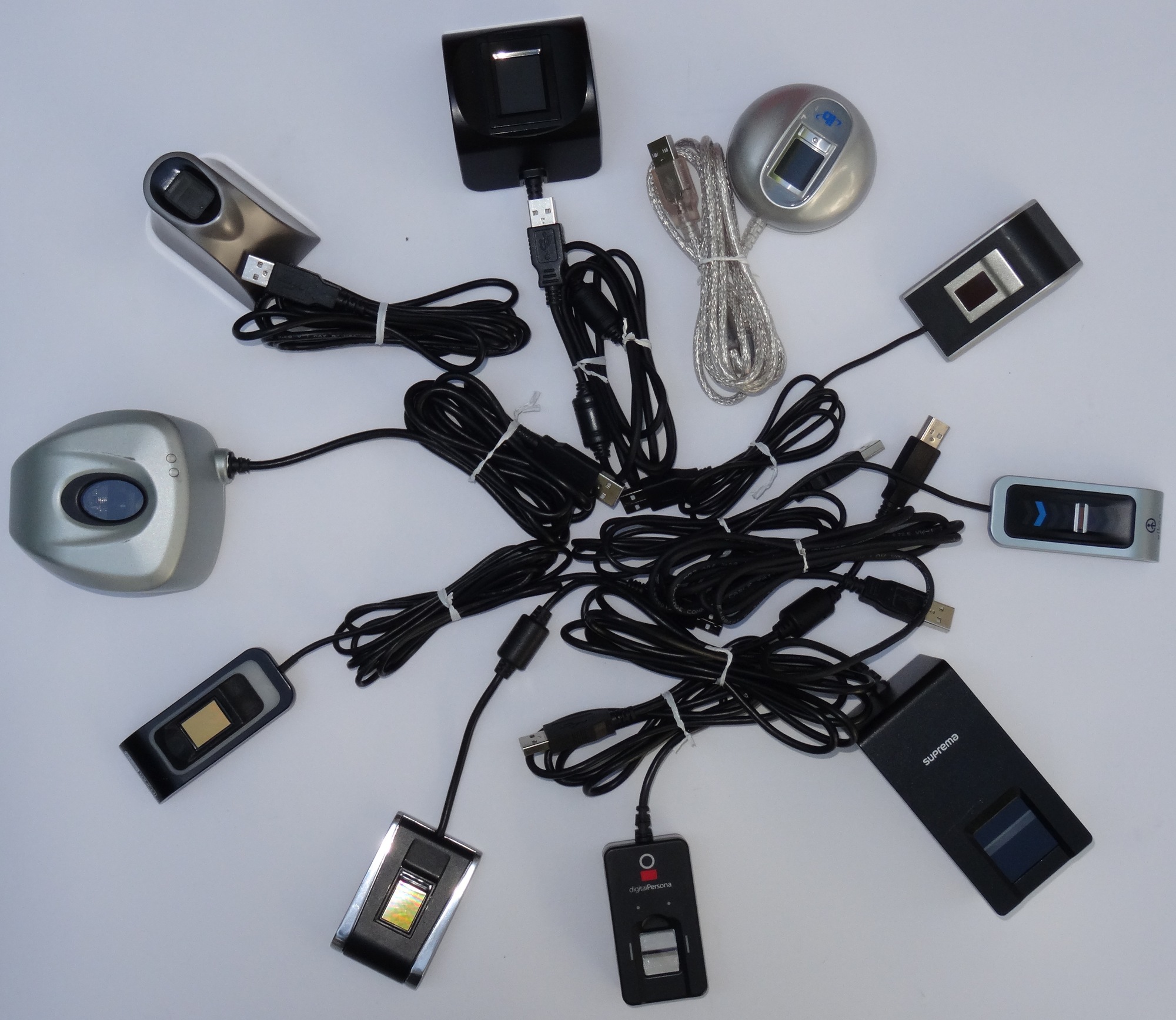}
\par\end{centering}
\caption{Fingerprint capturing devices, from middle left clockwise: Lumidigm
Venus V311 (optical multispectral), Lumidigm Mercury M311 (optical
multispectral), Integrated Biometrics Columbo (active capacitive),
Integrated Biometrics Curve (active capacitive), Next Biometrics NB-3010-U
(thermal), Upek Eikon II Swipe (capacitive swipe), Suprema Realscan
G1 (optical), DigitalPersona URU5160 (optical), Zvetco Verifi P5000
(capacitive), DigitalPerson Eikon 710 Touch (capacitive) \label{fig:Fingerprint-capturing-devices}}
\end{figure}

Prior to extracting the fingerprint features necessary to perform
the biometric recognition, a sample of the fingerprint has to be captures.
Such a sample can be acquired using different technologies of capturing
devices based on the optical properties of the skin surface, electrical
properties of the skin, thermal properties of the ridges and air gaps
of the valleys or ultrasonic reflectance properties of the skin \cite{BMaltoni03a}.
The first COTS fingerprint capturing devices appeared on the marked
in the 1980s and were mainly based on optical sensing technology.
Nowadays there is a shift towards capacitive technology and more recently
to ultrasonic devices, especially in recent smartphones as these sensors
are small enough to be built into virtually any appliance, while optical
devices are still predominant in high-security applications. Most
of the fingerprint capturing devices have a resolution of 500 dpi
at least. Some fingerprint capturing devices are shown in Figure \ref{fig:Fingerprint-capturing-devices}.
Each of those different technologies is described in the following.

\begin{figure}
\centering{}%
\begin{tabular}{cccc}
 &  &  & \tabularnewline
\includegraphics[height=0.07\paperheight]{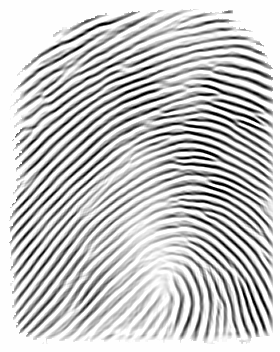} & \includegraphics[height=0.07\paperheight]{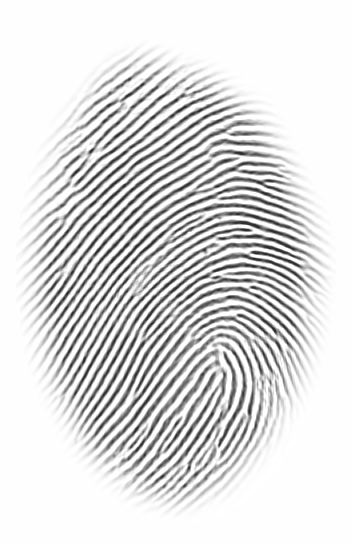} & \includegraphics[height=0.07\paperheight]{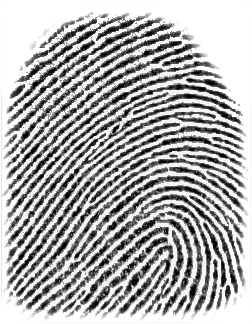} & \includegraphics[height=0.07\paperheight]{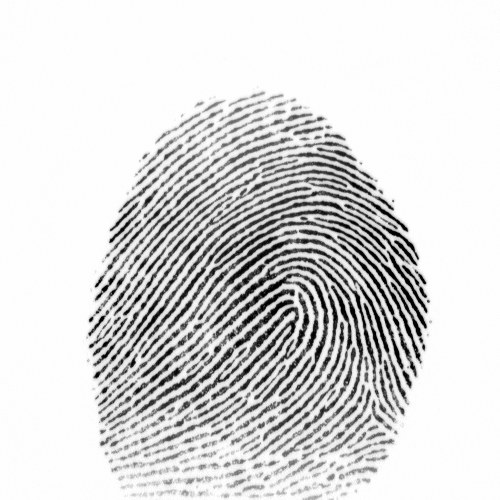}\tabularnewline
\includegraphics[height=0.07\paperheight]{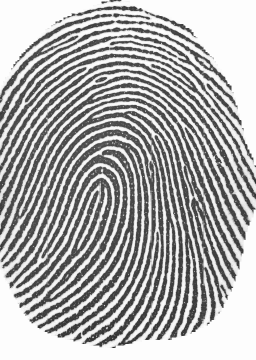} & \includegraphics[height=0.07\paperheight]{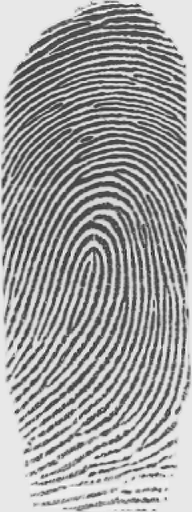} & \includegraphics[height=0.07\paperheight]{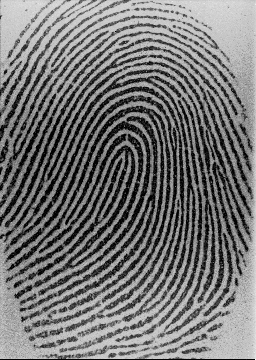} & \includegraphics[height=0.07\paperheight]{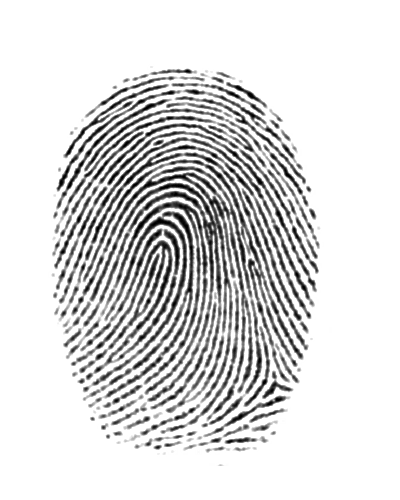}\tabularnewline
\includegraphics[height=0.07\paperheight]{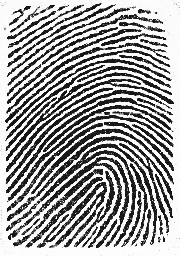} & \includegraphics[height=0.07\paperheight]{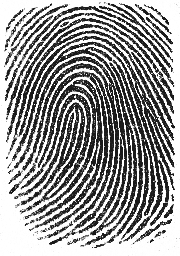} & \includegraphics[height=0.07\paperheight]{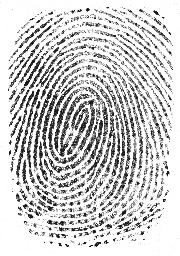} & \includegraphics[height=0.07\paperheight]{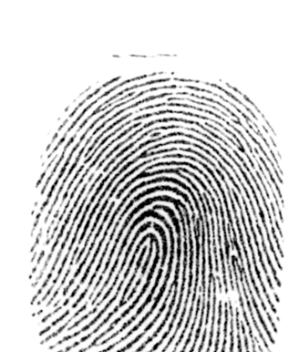}\tabularnewline
\end{tabular}\caption{Example fingerprint images, first row: optical, second row: capacitive,
third row: thermal (except the right one which is capacitive)}
\end{figure}

\paragraph{Optical devices}

are the most widely used ones, followed by capacitive devices. The
finger is placed on a glass surface, which is essentially a transparent
prism. An optical fingerprint capturing device contains some kind
of light source to illuminate the fingertip surface (usually based
on light emitting diodes) and a camera to capture the image. Most
optical devices use the total internal reflection (TIR) principle.
The fingertip is placed directly on the glass plate with the ridges
directly touching the glass plate. The surface is illuminated by the
internal light source through one side of the prism. The light entering
the prism is reflected at the valleys and absorbed at the ridges,
creating a contrast between ridges and valleys. The reflected light
is projected on the camera through a system of lenses and captured
by the camera. Optical devices are sensitive to spoofs that are composed
of materials exhibiting the same light reflectance as human skin.
A special kind of optical devices are multispectral capturing devices
which utilise more than one light source or a light source capable
of emitting different wavelengths of light. This should improve the
resistance to presentation attacks. More recently also contactless
optical fingerprint capturing devices entered the market. Those devices
essentially consist of an image sensor and an axillary light source.
They do not require direct contact between the fingertip and the sensor,
which eliminates hygienic issues at the one hand and problems with
dirt and soiling of the sensor surface on the other hand.

\paragraph{Electric field devices}

also called solid state devices or active capacitive sensors. These
devices consist of an array of pixels able to measure differences
in the electric field. They create an electric field between the sensor
surface and the finger surface. Due to variations in the conductive
layer of the skin, which is located beneath the skin surface, there
are variations in the electric field of the ridges and valleys, which
are then measured and translated into a digital representation (i.e.
an output image). Those sensors have kind of a metal or conductive
point (usually a bar at the bottom of the sensor surface) which has
to be touched in order for the sensor to create the electric field. 

\paragraph{Capacitive devices}

also known as passive capacitive sensors. As the name suggests, those
devices exploit the fact that there is a difference in the capacity
between the ridges and valleys of the fingertip as the ridges directly
touch the sensor surface, while there is a small air gap between the
sensor surface and the valleys. This difference in the capacitance
values is measured. The finger surface acts as one electrode of a
capacitor with the sensor surface as the second electrode. The whole
sensor surface consists of capacitance sensitive ``pixels''. Capacitive
sensors can be built very small and even integrated into smart cards.
They are more vulnerable to gelatin based fake fingerprint representations.

\paragraph{Thermal devices}

are based on a combination of silicon and a pyroelectric material,
which measures differences in temperature and converts these into
a digital output signal. When the fingertip is placed at the sensor,
the ridges directly touch the sensor surface, changing the sensor
surface temperature, while the valleys are isolated by a small air
gap between the skin and the sensor surface, hence the temperature
remains constant where the valleys are located. This creates the necessary
difference in temperatures between ridges and valleys which can then
be measures. A disadvantage is that after some time the image disappears
as the sensor reaches a thermal equilibrium, i.e. there is a constant
temperature all over the sensor surface. In order to reduce production
costs and device dimensions (smaller sensor area), the first thermal
devices were built as so called swipe sensors, i.e. sensor surface
does not cover the whole fingertip area but only a small bar with
the same width as the width of the fingertip but only a few pixels
in height. The finger is slid over the sensor surface (in a swiping
motion) and the sensor generates the output image by stitching together
small image patches of the different regions of the fingertip during
the sliding process. In general, swipe sensors are considered as less
user-friendly due to the necessary swiping motion. Thermal sensors
are rarely used nowadays.

\paragraph{Ultrasound devices}

These devices exploit the difference in the acoustic impedance between
the skin of the ridges and the air of the valleys (the ridges directly
touch the sensor surface). The sensor itself is an acoustic transmitter,
which transmits an acoustic signal towards the fingertip surface and
captures the signal which is received. The frequency range of those
devices is from about 20 kHz to several GHz. Higher frequencies correspond
to higher resolution of the resulting images. Ultrasound devices used
to be more expensive than other fingerprint capturing devices but
with the advent of new fabrication technologies and in-screen sensing
solutions for smartphones, those devices became more popular.

\section{Literature Review on Fingerprint Presentation Attacks\label{sec:Literature-Review}}

There are several possibilities to launch a presentation attack or
to spoof a fingerprint scanner. All of them have in common that someone
wants to trick the system, either in order to gain illegitimate access
to it or to impersonate another person for various reasons (e.g. to
have a shared personality). For a presentation attack to be successful,
not only the spoofing detection of the fingerprint scanner has to
be circumvented/tricked, but also during the matching process, the
match score has to be high enough to allow access to be granted. Besides
creating a spoofing artefact, other options include: forcing the legitimate
user to present his finger to the scanner device, presenting a cadaver
finger to the device or use the latent imprint left on the scanner
device. As we focus on physical fake fingerprint representation (spoofing
artefacts), the aforementioned other possibilities are out of scope
of this work. 

Creating a physical fake fingerprint representation usually involves
two steps: creation of the mould (negative imprint) and creating of
the cast. At first, either the legitimate user presses his finger
on the mould material (cooperative duplication), or some kind of latent
representation is used for that (non-cooperative duplication). Afterwards,
depending on the mould it needs some time to cure. Afterwards the
cast material is filled in the mould, let there to cure and removed
afterwards, resulting in the final fake fingerprint representation.
Table \ref{tab:overview_mould} listing common mould and cast materials
and techniques. Besides the mould-cast method, there are also some
materials for which no mould is needed, i.e. the genuine fingerprint
is directly applied on the cast material. These materials are listed
in Tab. \ref{tab:overview_nomould}. In the following, the cooperative
and non-cooperative duplication are described in more detail.

\begin{table*}[htb]
	\centering
	\caption{with mould}
	\label{tab:overview_mould}
	\begin{tabular}{@{}l p{2cm} p{2.5cm} p{4cm} p{3cm}@{}}
	\toprule
	Reference & Method & Mould Material & Artefact Material & Sensor Type \\
	\toprule
	Van der Putte and Keuning \cite{BVanDerPutte00a} & cooperative,\newline non-cooperative & Plaster\newline Printed circuit board & Silicone cement & n/a\\ 	\midrule
	Schuckers \cite{BSchuckers02a} & cooperative & Dental impression material & Play-Doh\newline Clay & Desktop: capacitive, optical \\
	\midrule
	Matsumoto \cite{BMatsumoto02a} & cooperative,\newline non-cooperative & Moulding plastic & Gelatine\newline Silicone + conductive ink & Desktop: optical, capacitive \\
	\midrule
	St{\'e}n, Kaseva and Virtanen \cite{BSten03a} & cooperative,\newline non-cooperative & Hot glue\newline Printed circuit board & Grease + Breath\newline Gelatine & Desktop: Capacitive\\
	\midrule
	Galbally \etal \cite{BGalbally06a} & cooperative & Modeling putty & Modeling silicone & Desktop: optical, thermal swipe\\
	\midrule
	Barral and Tria \cite{BBarral09a} & cooperative & Wax\newline Silicone\newline FIMO paste & Glycerine & Desktop: optical, capacitive touch, capacitive swipe, thermal swipe\\
	\midrule
	Espinoza and Champod \cite{BEspinoza11a} & cooperative,\newline non-cooperative & Utile plast\newline Siligum & Latex\newline White glue & Desktop: multispectral\\
	\midrule
	Espinoza, Champod and Margot \cite{BEspinoza11b} & cooperative,\newline non-cooperative & Utile plast\newline Siligum\newline Printed acetate sheet & Latex\newline White glue & Desktop: optical\\
	\midrule
	Rattani and Ross \cite{BRattani14a} & cooperative & Wax\newline Play-Doh\newline Plaster & Gelatine\newline Silicone\newline Wood glue\newline Ecoflex\newline Latex & Desktop: optical \\
	\midrule
	Cao and Jain \cite{BCao16a} & non-cooperative & Conductive ink & Latex\newline Wood glue & Smartphone: Touch\\
	\midrule
	Gonzalo \etal \cite{BGonzalo18a} & cooperative & BluTack\newline Candle wax\newline Hot glue\newline Plasticine\newline Play-Doh\newline Siligum\newline Stamp wax\newline Unbranded silicon & Alginate\newline Art glue\newline Art glue + graphite\newline BluTack\newline Body wax + conductive paint\newline Candle wax (+ conductive ink)\newline Facemask\newline Gelatine powder (+ conductive ink)\newline Gelatine powder + glycerine\newline Gelatine sheet + (graphite/water)\newline Plasticine (+ conductive ink)\newline Play-Doh & Smartphone: Swipe, Touch\\
	\midrule
	Kanich, Drahansky and M{\'e}zl \cite{BKanich18a} & cooperative,\newline semi-cooperative & Printed circuit board & Fimo standard\newline Fimo Air\newline Kera\newline Hobby Mass\newline Magic putty\newline WePAM\newline Mamut glue\newline Acrylic sealant\newline Herkules glue\newline Oyumare\newline Play-Doh\newline Vegetable play-doh\newline Premo\newline Tropicalgin\newline Glass colors\newline Cernit\newline Gel wax\newline Kato\newline Siligum\newline Latex\newline Wax sheets & Desktop: optical, multispectral, pressure sensitive \\
	\bottomrule
 	\end{tabular}
\end{table*}

\begin{table*}[htb]
	\centering
	\caption{no mould}
	\label{tab:overview_nomould}
	\begin{tabular}{@{}l p{2cm} p{3.5cm} p{5cm}@{}}
	\toprule
	Reference & Method & Artefact Material & Sensor Type \\
	\toprule
	Goicoechea-Telleria \etal \cite{BGoicoecheaTelleria18a} & cooperative,\newline non-cooperative & Play-Doh\newline Gelatine\newline Clay\newline Wood glue\newline Conductive ink\newline Latex\newline Latex + Graphite\newline Silicone\newline Silicone + graphite & Desktop: Thermal, Capacitive, Optical\newline Smartphone: Swipe, Touch \\
	\midrule
	Marcialis \etal \cite{BLivDet2009} (LivDet09) & cooperative  & Play-Doh\newline Silicone \newline Gelatine & Desktop: optical\\
	\midrule
	Yambay \etal \cite{BLivDet2011} (LivDet11) & cooperative  & Play-Doh\newline Silicone \newline Gelatine\newline Latex\newline Wood glue & Desktop: optical\\ 	\midrule
	Orr{\`u} \etal \cite{BLivDet2019} (LivDet19) & cooperative & Mix1\newline Mix2\newline Liquid Ecoflex\newline Ecoflex\newline Body double \newline Gelatine\newline Latex\newline Wood glue & Desktop: optical, thermal swipe\\
	\bottomrule
 	\end{tabular}
\end{table*}

\subsection{Cooperative Duplication}

The first variant of creating an artificial (fake) fingerprint is
cooperative duplication. This involves the active participation of
the live subject (fingerprint donor). The finger of the subject is
at first pressed on the surface of the mould material, usually some
kind of cast, dental impression material, plaster or modelling clay.
Once the negative impression of the finger is fixed on it, the finger
is removed and depending on the material, the mould is cured in an
oven or at room temperature for some time. The actual spoof (cast)
is then created by filling the mould with some liquid material (e.g.
silicone, latex, gelatin or glue) and left in the mould until it has
hardened. Once removed, the fake fingerprint (spoof) is created. Usually
spoofs created by cooperative duplication are of higher quality than
the ones created by non-cooperative duplication.

\subsection{Non-Cooperative Duplication}

There are several ways to create a fake fingerprint without the cooperation
of the genuine subject. The first one are latent fingerprints, left
on a surface by the subject. These latent prints can be lifted with
powder and then covered with Scotch tape to lift the print. This print
can directly be used as a fake fingerprint representation. Another
method is to use a PCB (printed circuit board) to enhance the quality
of the fake fingerprint. At first the fingerprint is placed on a transparency
and brushed with black powder. Afterwards a digital photograph of
the fingerprint is taken to create a mask, which is then placed on
the PCB and etched using UV light. This PCB imprint now serves as
a mould which can now be filled with some liquid material in the same
way as the moulds in cooperative duplication. Another ways of non-cooperative
duplication are fingerprint reactivation, where a latent print on
the sensor itself is reactivated, e.g. by breathing or placing a water-filled
plastic bag on the sensor surface in order to fool the sensor. Using
a cadaver fingerprint and fingerprint synthesis (creating a synthetic
fingerprint based on a biometric template) can also be utilized to
acquire a fake fingerprint in an non-cooperative way. 

In the rest of this work we focus on cooperative duplication only.
All of the tested moulds and cast materials involve cooperative duplication.

\section{Experimental Evaluation\label{sec:Experimental-Evaluation}}

In the following the evaluation methodology and experimental protocol
for fabricating and testing the spoofed fingerprint representations
are described. Afterwards the fabrication of the different moulds
is described in detail, followed by the fabrication of the different
casts. Finally, the evaluation results in terms of fingerprint quality
as well as matching scores are given and discussed.

\subsection{Evaluation Protocol and Methodology}

During our experiments we utilised five COTS fingerprint scanner devices
to cover most of the available sensing technologies, including optical,
thermal as well as capacitive:
\begin{itemize}
\item Lumidigm Venus V311 (optical, multispectral) \cite{LumidigmV311Website}
\item Suprema RealScan G1 (optical) \cite{SupremaG1Website}
\item Next Biometrics NB-3010-U (thermal) \cite{NB-3010-UWebsite}
\item Integrated Biometrics Columbo (active capacitive) \cite{IBColumboWebsite}
\item Zvetco Verifi P5000 (passive capacitive) \cite{VerifiP5000Website}
\end{itemize}
To evaluate the effectiveness of the fabricated spoof fingerprints,
we used a two step protocol. At first we fabricated a small number
of moulds and casts for each type of material. The casts were presented
to the fingerprint scanners and the result in terms of fingerprint
quality, evaluated using Neurotechnology VeriFinger's (Version 10.0)
\cite{VeriFingerWebsite} built-in fingerprint quality assessment
was recorded. If the scanner did not react at all while presenting
the spoofed representation or if it detected it as a spoof, there
is no quality value recorded.

Afterwards, the five best working materials/spoof types in terms of
fingerprint quality were selected for the production of a larger number
of spoofed fingerprints from 15 different subjects. These spoofs were
presented to the fingerprint scanners and enrolled in the database
(the respective template was created and stored). Afterwards, genuine
impressions of the subjects' fingers were enrolled as well. By comparing
the templates created from the spoofed representation to the ones
created from the genuine subject's real finger (bona fide), a quantification
of the spoofed representations effectiveness can be done based on
the matching scores.

\subsection{Fabrication of the Moulds\label{subsec:Molds}}

Several different types of moulds were created tested which are described
in the following:

\paragraph{Cast: }

\begin{figure}
\begin{centering}
\includegraphics[height=6cm]{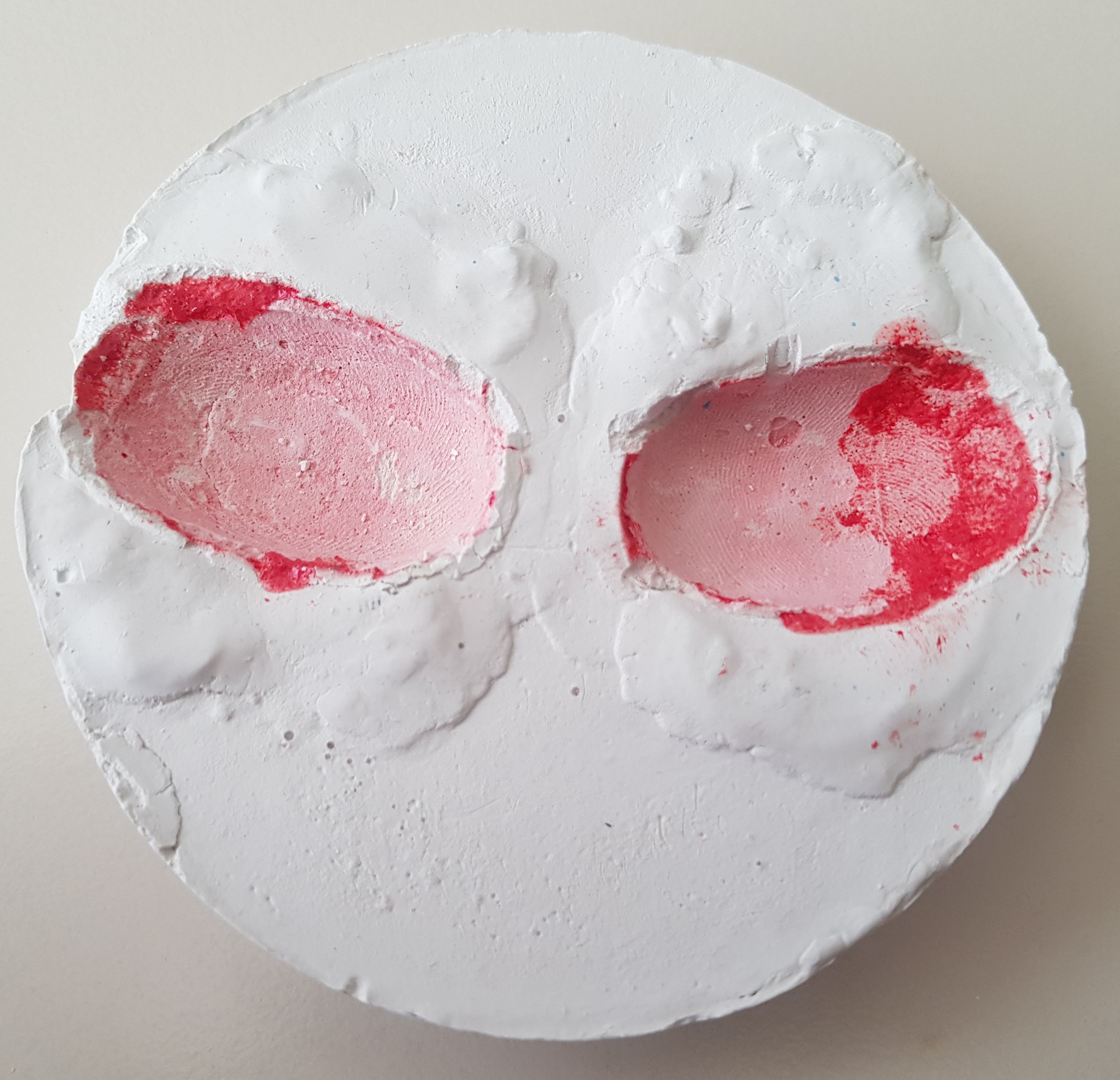}
\par\end{centering}
\caption{Mould made from cast\label{fig:Mold-made-from-cast}}
\end{figure}

For creating this mould, modelling cast from Knauf was used. The cast
was first mixed with tap water. Then it was let for curing about 6
min. After that the finger was put and pressed into the clay for about
3 min. Then it was let for curing about 30 min until it hardened.
As an alternative, it was first let for curing 10 min instead of 6
min, which creates a flat impression of the fingerprint instead of
a deep one. An example image can be seen in Fig. \ref{fig:Mold-made-from-cast}.

\paragraph{Candle wax: }

\begin{figure}
\begin{centering}
\includegraphics[height=6cm]{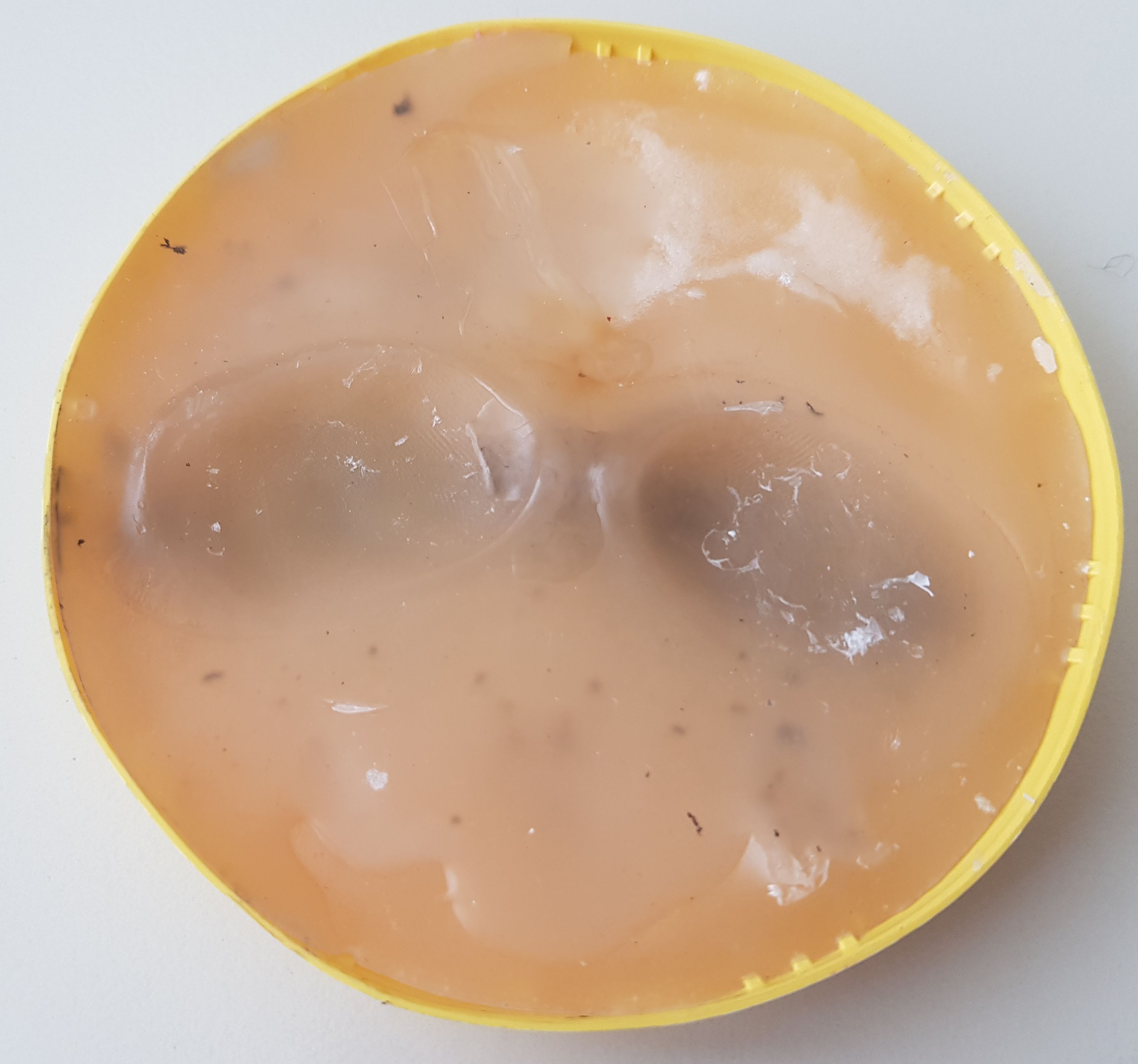}\hspace{0.05\paperwidth}\includegraphics[height=6cm]{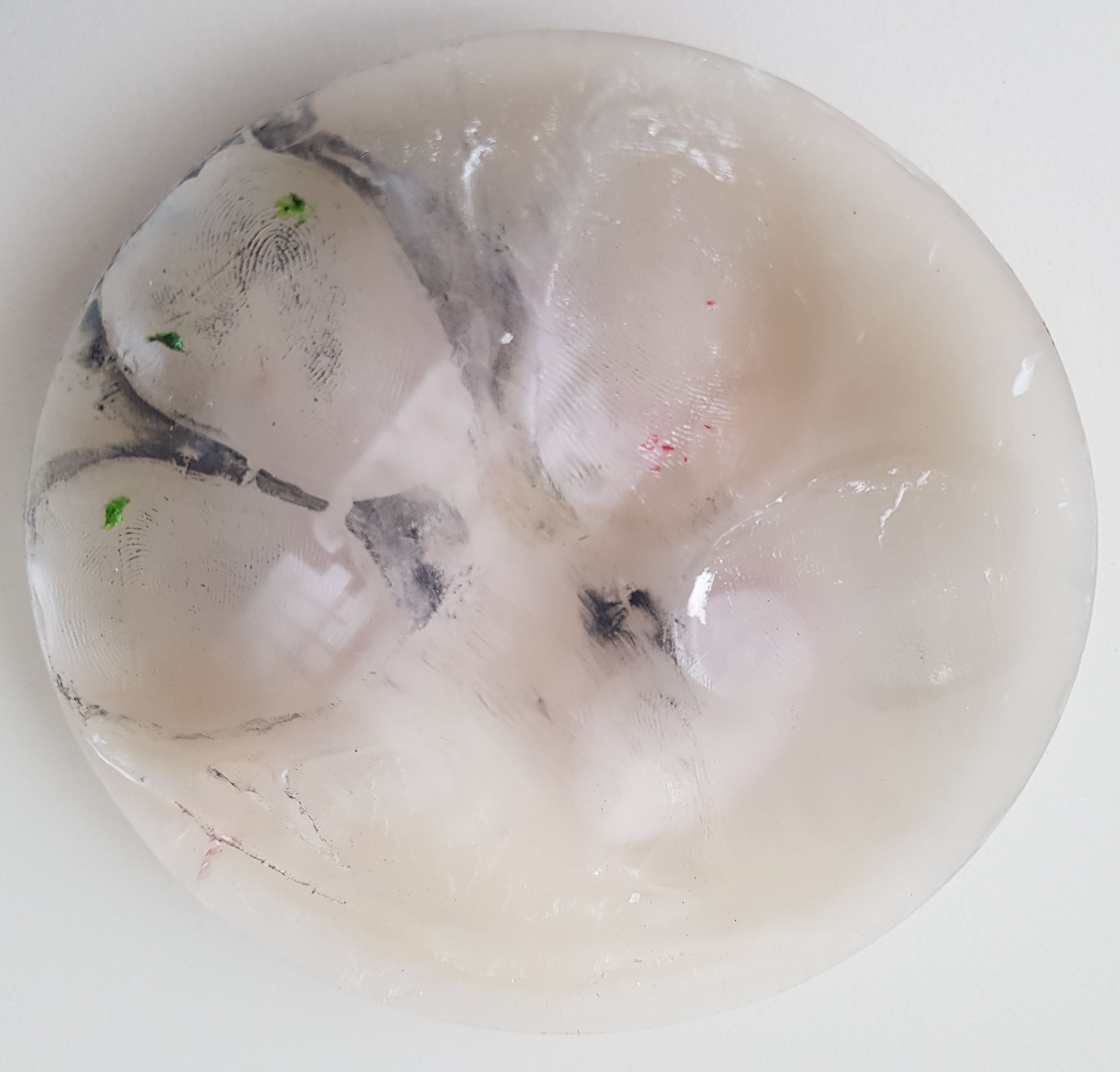}
\par\end{centering}
\caption{Mould made from candle wax\label{fig:Mold-made-from-candlewax}}
\end{figure}

Wax was heated in a small glass bowl on an electric cooking plate
at first (1 min on level 5, then about an additional minute on level
2 until it was completely molten). Afterwards we let it cool down
in the glass bowl for about 5 min, then the finger is pressed into
the semi hardened wax. Some candle wax moulds are depicted in Fig.
\ref{fig:Mold-made-from-candlewax}

\paragraph{Acrylic: }

The acrylic was put in a small glass bowl as well and let there for
curing about 2h. Afterwards the finger was pressed into the acrylic
for about 1 min to create the imprint. With the acrylic we were not
able to produce good imprints as either it sticked to the finger or
it was to dry to create a good impression - too many friction ridges
and no smooth surface. Hence, this type of mould was not used during
the subsequent tests.

\paragraph{Silicone: }

Same procedure as with the acrylic. Silicone was put into a small
glass bowl and cured for about 2h. Afterwards the finger was pressed
into it. Again, no good impression due to friction ridges and non-smooth
surface. This type of mould was not used during the subsequent tests.

\paragraph{Hair wax: }

The wax was prepared in a bowl and mixed with window colour (otherwise
it would be transparent) and the finger was pressed into it immediately
after pouring it into the bowl. Due to the high moisture of the hair
wax, no good impression was possible and the window colour did not
cure/dry properly. This type of mould was not used during the subsequent
tests.

\paragraph{Play-Doh (putty):}

At first some Play-Doh was prepared and spread on a surface. Afterwards
the finger is pressed into the clay to create the impression. By applying
varying pressure, the depth of the imprint can be varied (from flat
to deep).

\paragraph{Siligum: }

\begin{figure}
\begin{centering}
\includegraphics[height=6cm]{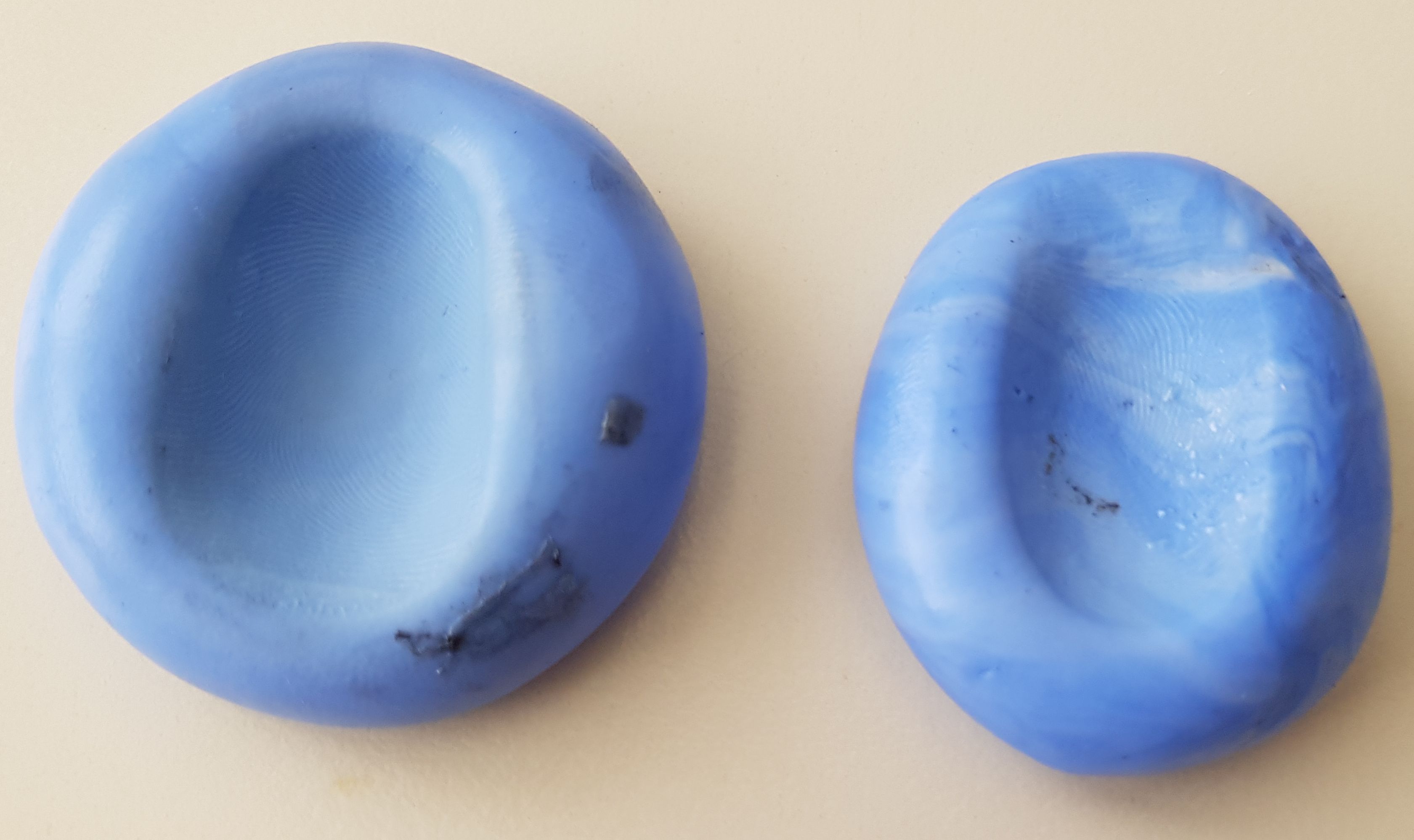}
\par\end{centering}
\caption{Mould made from Siligum\label{fig:Mold-made-from-siligum}}
\end{figure}

Siligum is a dental cast material. To create the mould, a round block
of Siligum was formed, the finger was pressed into the mould and then
the mould was left at room temperature for an hour to cure and harden.
Example Siligum moulds are shown in Fig. \ref{fig:Mold-made-from-siligum}.

\paragraph{Fimo: }

\begin{figure}
\begin{centering}
\includegraphics[height=6cm]{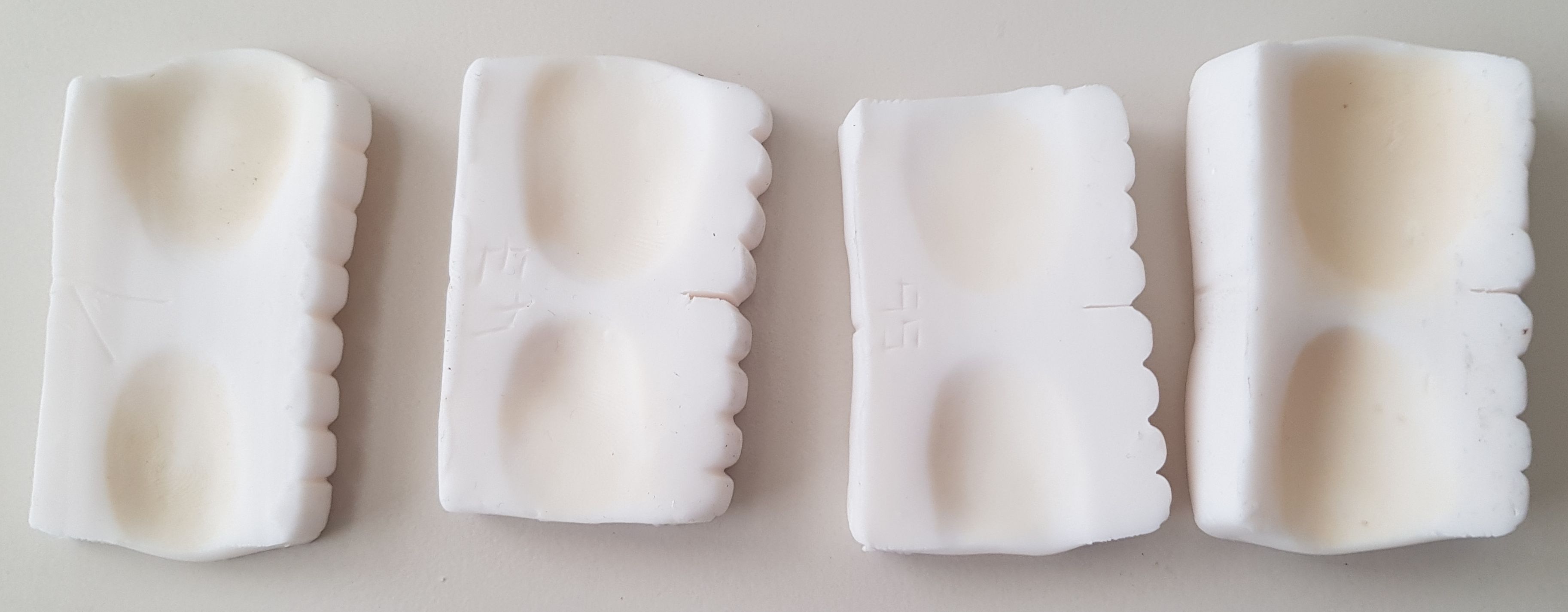}
\par\end{centering}
\caption{Mould made from Fimo\label{fig:Mold-made-from-fimo}}
\end{figure}

Fimo is a modelling clay which is available in different variants.
Some cure at room temperature, some need to be put in the oven for
curing. We used the latter one. At first the finger was pressed into
a block of Fimo, using the thumb of the other hand to increase the
pressure. Afterwards, the block was put in the oven at 110\textdegree C
for 30min and then left at room temperature for another 2 hours to
cure. Several Fimo moulds are depicted in Fig. \ref{fig:Mold-made-from-fimo}.
The Fimo moulds were easier to produce and worked better than the
cast and wax ones, thus we decided for Fimo and Cernit moulds to produce
the majority of the casts.

\paragraph{Cernit: }

\begin{figure}
\begin{centering}
\includegraphics[height=6cm]{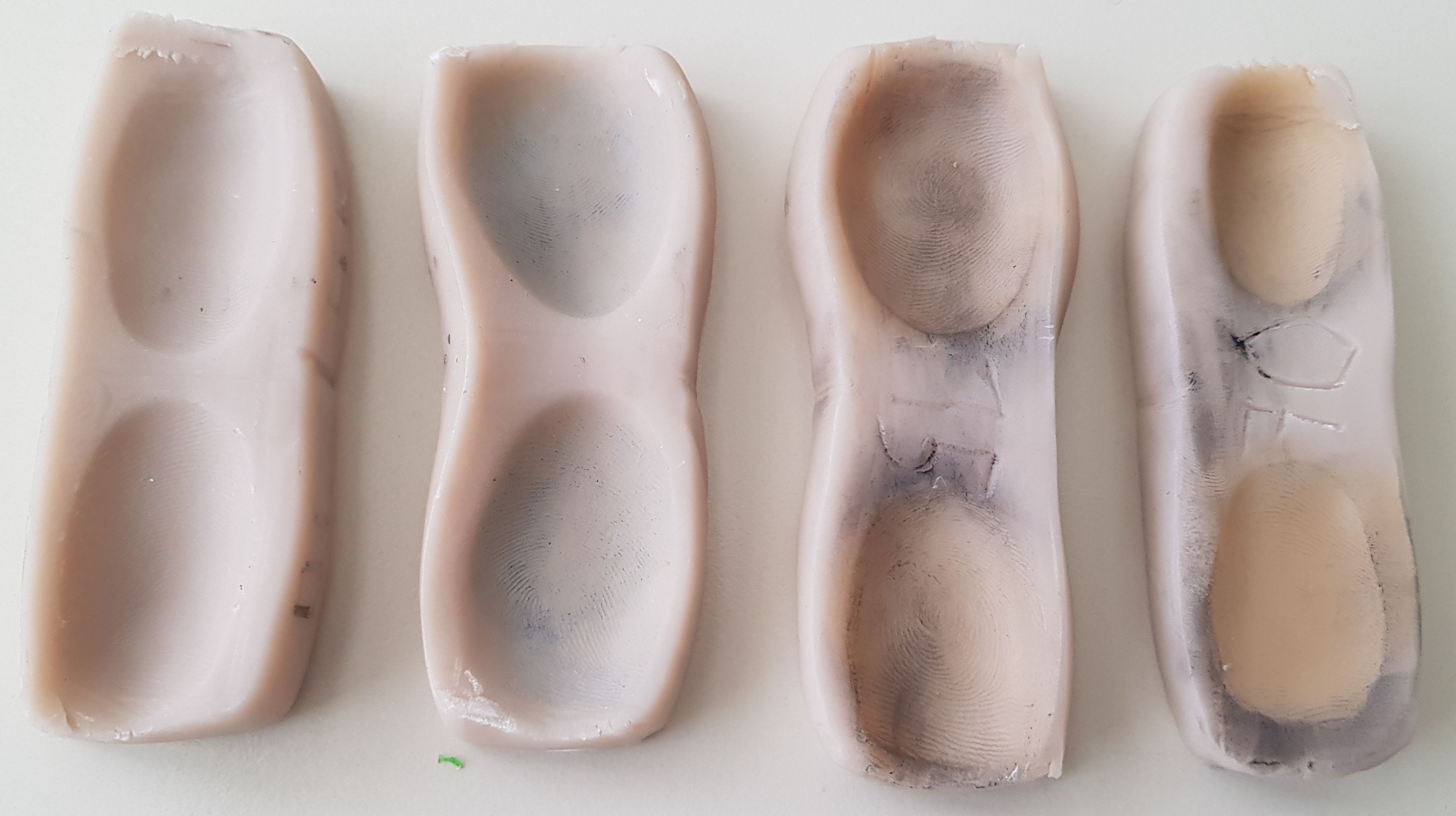}
\par\end{centering}
\caption{Mould made from Cernit\label{fig:Mold-made-from-cernit}}
\end{figure}

Cernit is basically the same type of modelling clay as Fimo, just
a different manufacturer. Again to create the mould, the finger was
pressed into a block of Cernit, with the thumb of the other hand to
apply additional pressure. Then the block was put in the oven for
30min at 110\textdegree C and left at room temperature for another
2 hours to cure. A few Cernit moulds can be seen in Fig. \ref{fig:Mold-made-from-cernit}.
For most subsequent casts (except for the initial tests), either Fimo
or Cernit moulds have been used.

\subsection{Fabrication of the Casts}

The casts were created by pouring the cast material into one of the
moulds. For most types of casts, either the cast moulds or the Cernit/Fimo
moulds were used. Fig. \ref{fig:Overview-casts} shows examples of
all the different casts that were used during the experiments. Each
attempt (cast/mould combination) has a number, which is later used
in subsection \ref{subsec:Spoofed-Fingerprint-Quality} and \ref{subsec:Spoofed-Fingerprint-Matching}
to identify the results. The different cast materials and their fabrication
are described in the following:

\begin{figure}
\begin{centering}
\includegraphics[width=1\textwidth]{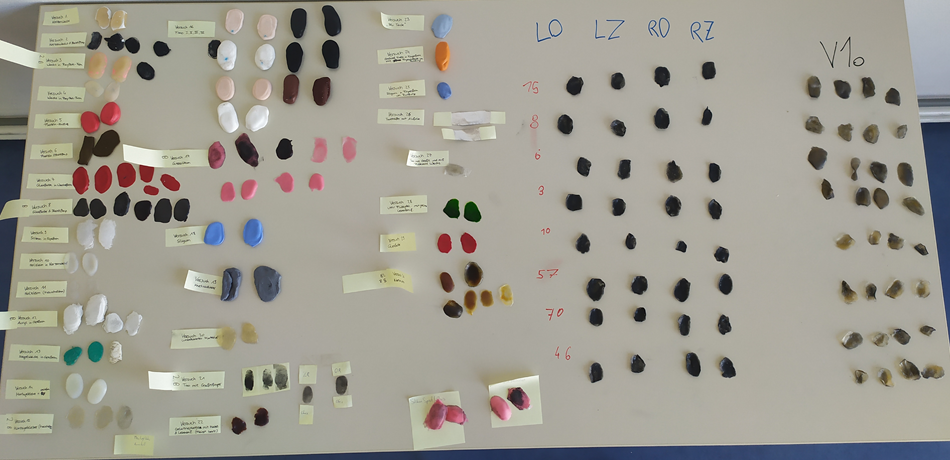}
\par\end{centering}
\caption{Overview of the casts\label{fig:Overview-casts}}
\end{figure}

\paragraph{Candle Wax:}

\begin{figure}
\begin{centering}
\includegraphics[height=4cm]{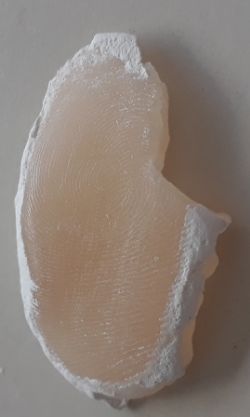}\hspace{0.05\paperwidth}\includegraphics[height=4cm]{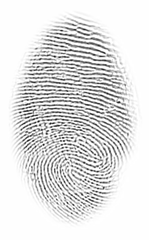}\hspace{0.05\paperwidth}\includegraphics[height=4cm]{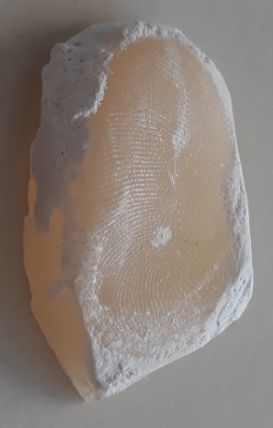}
\par\end{centering}
\caption{Candle wax casts\label{fig:Casts-candlewax}}
\end{figure}

(\textit{Attempt \#1}) At first the candle wax is poured in a small
glass bowl and heated on the electric cooking plate, same as for the
mould, at first 1min on level 5, then an additional minute on level
2 until it is completely melted. Afterwards it is cooled down for
5 min and then poured into the cast mould, left there for about 30min
to cool down and cure. Finally the cast is carefully removed from
the mould. Fig. \ref{fig:Casts-candlewax} shows some example images
of the candle wax mould. It can be seen that some parts of the cast
mould still stick to the candle wax cast. The candle wax cast is rather
hard and inflexible, making it difficult to apply it to the fingerprint
capturing devices without breaking it. It worked with the Lumidigm
scanner, but all other scanners showed no reaction at all.

\begin{figure}
\begin{centering}
\includegraphics[height=4cm]{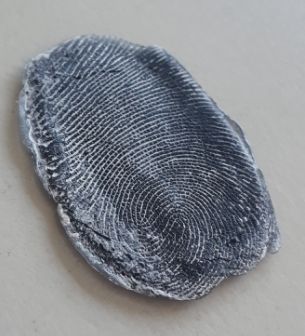}\hspace{0.05\paperwidth}\hspace{0.05\paperwidth}\includegraphics[height=4cm]{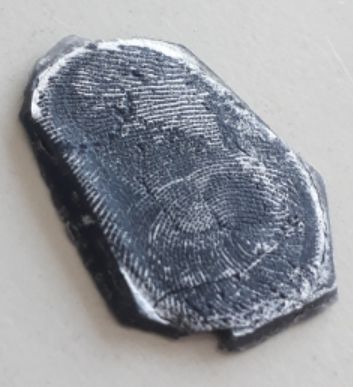}
\par\end{centering}
\caption{Candle wax casts with graphite powder\label{fig:Casts-candlewax-graphite}}
\end{figure}

The first approach with candle wax did not work well with the tested
fingerprint capturing devices. Hence we did a second try by mixing
the liquid candle wax with graphite powder to increase the contrast
and conductivity. Again, the cast is rather solid and does not work
for most fingerprint readers. Example images of this second approach
are depicted in Fig. \ref{fig:Casts-candlewax-graphite}. This approach
made the results worse as now none of the fingerprint scanners showed
any reaction.

The third attempt (\textit{Attempt \#2}) was to spread graphite powder
inside the cast mould and then pouring the liquid candle wax into
the mould in order to overcome the difficulties with removing the
cast from the mould. However, this did not improve the situation,
the cast was still difficult to remove from the mould and the results
were comparable to the second attempt with the graphite powder mixture.
Again, these casts only worked with the Lumidigm scanner.

\begin{figure}
\begin{centering}
\includegraphics[height=4cm]{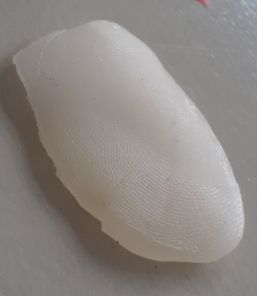}\hspace{0.05\paperwidth}\includegraphics[height=4cm]{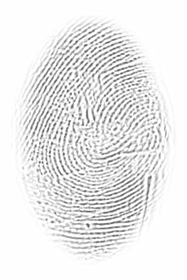}\hspace{0.05\paperwidth}\includegraphics[height=4cm]{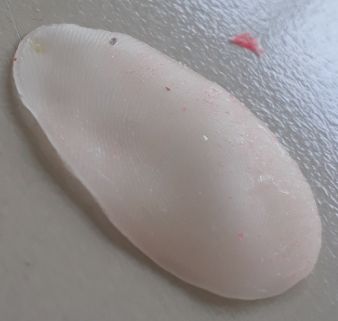}
\par\end{centering}
\caption{Candle wax casts in Play-Doh mould\label{fig:Casts-candlewax-playdoh}}
\end{figure}

The last attempt (\textit{Attempt \#3 and \#4}) was to use the Play-Doh
mould instead of the cast one. The liquid candle wax was poured into
the Play-Doh mould and left there to cure for about 30min. Afterwards
the cast was removed from the mould. This time, the removal procedure
was much easier than with the cast mould and also the results with
the fingerprint capturing devices were more promising, at least with
the Lumidigm sensor. Fig. \ref{fig:Casts-candlewax-playdoh} shows
some examples of this kind of cast. Again, this attempt only worked
with the Lumidigm scanner while all others still showed no reaction
at all.

\paragraph{Play-Doh:}

\begin{figure}
\begin{centering}
\includegraphics[height=4cm]{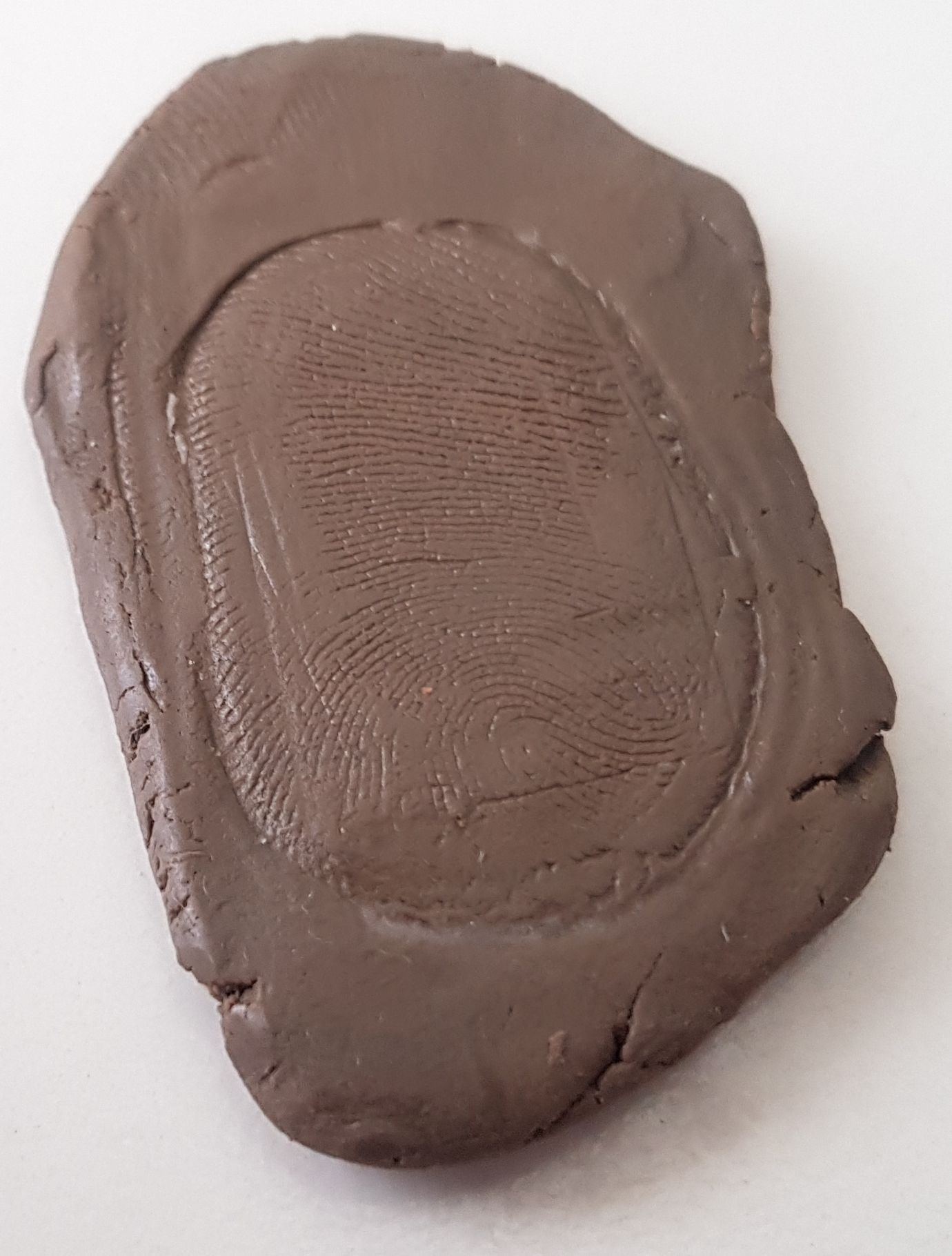}\hspace{0.05\paperwidth}\includegraphics[height=4cm]{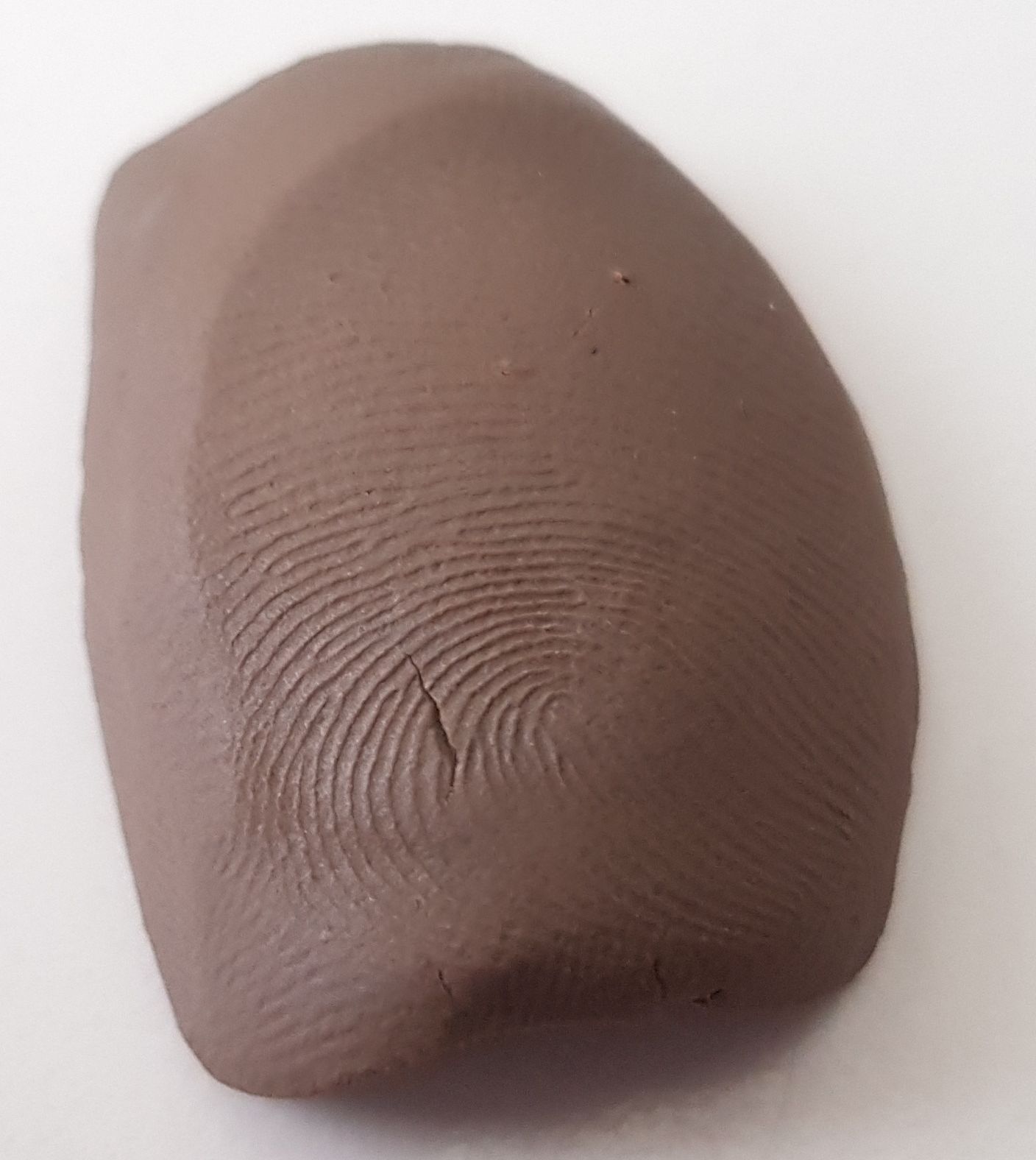}
\par\end{centering}
\caption{Play-Doh casts\label{fig:Casts-playdoh}}
\end{figure}

The Play-Doh putty was put into the cast mould, left there for about
1 minute and then removed. Unfortunately none of the fingerprint capturing
devices showed any reaction when the cast was presented to the sensor.
None of the tested fingerprint scanners showed any reaction when the
Play-Doh casts were presented to the scanner. Some Play-Doh casts
are shown in Fig. \ref{fig:Casts-playdoh}.

\paragraph{Plasticine:}

\begin{figure}
\begin{centering}
\includegraphics[height=4cm]{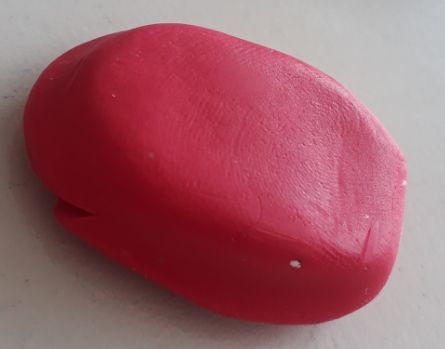}\hspace{0.05\paperwidth}\includegraphics[height=4cm]{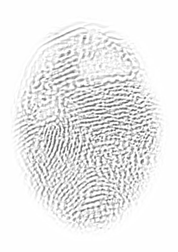}\hspace{0.05\paperwidth}\includegraphics[height=4cm]{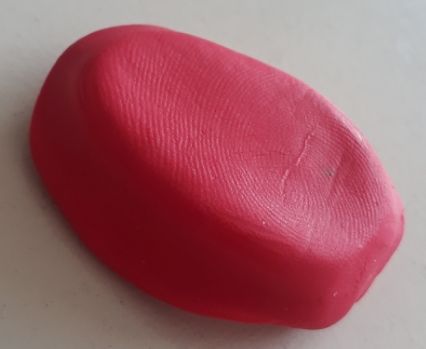}
\par\end{centering}
\caption{Plasticine casts\label{fig:Casts-plasticiline}}
\end{figure}

(\textit{Attempt \#5}) Red plasticine putty was put into the cast
mould. The cast worked for the Lumidigm scanner but did not work for
the other tested devices. Another problem with this type of cast is
that it wears off over time and with each use due to the pressure
that needs to be applied while presenting the cast to the fingerprint
reader. Fig. \ref{fig:Casts-plasticiline} depicts some example images.
The plasticine casts worked with the Lumidigm scanner. The RealScan
scanner detected them as spoofed fingerprints. All other scanners
showed no reaction at all.

\begin{figure}
\begin{centering}
\includegraphics[height=4cm]{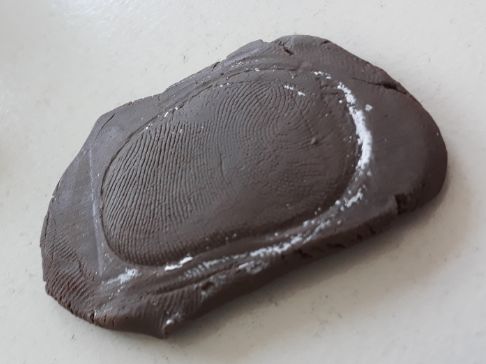}\hspace{0.05\paperwidth}\includegraphics[height=4cm]{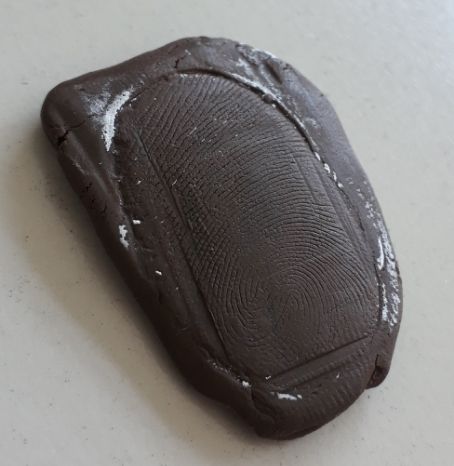}\hspace{0.05\paperwidth}\includegraphics[height=4cm]{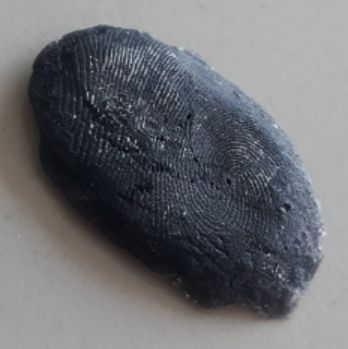}
\par\end{centering}
\caption{Plasticine casts with graphite powder\label{fig:Casts-plasticiline-graphite}}
\end{figure}

The second variant (\textit{Attempt \#6}) of the plasticine cast was
to mix plasticine with graphite powder in order to increase the conductivity
for the capacitive sensors. These casts are shown in Fig. \ref{fig:Casts-plasticiline-graphite}
and worked for the passive capacitive fingerprint reader but not for
the active capacitive one.

\paragraph{Window Colour:}

\begin{figure}
\begin{centering}
\includegraphics[height=4cm]{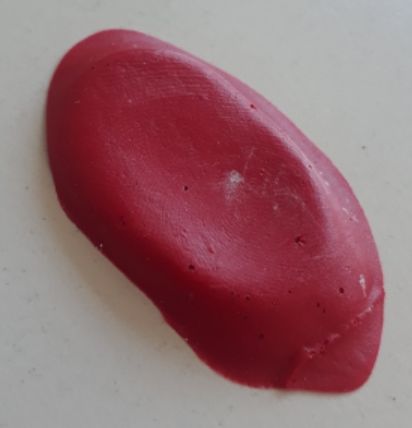}\hspace{0.05\paperwidth}\includegraphics[height=4cm]{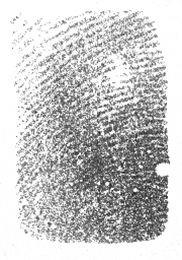}\hspace{0.05\paperwidth}\includegraphics[height=4cm]{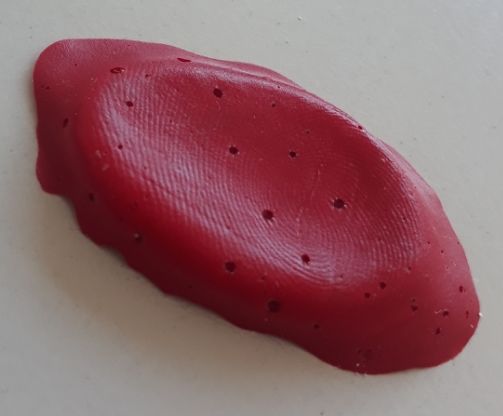}
\par\end{centering}
\caption{Window colour casts\label{fig:Casts-window-colour}}
\end{figure}

(\textit{Attempt \#7 and \#29}) Red window colour was used in combination
with the candle wax mould. It was poured into the mould and left there
for 5 days to cure. Interestingly this type of cast worked with the
thermal fingerprint scanner bit with none of the other fingerprint
scanners. A problem with this type of cast is again that it is a rather
soft material which wears of rather quickly due to the pressure applied
while presenting it to the fingerprint scanner. Some examples of this
cast can be seen in Fig. \ref{fig:Casts-window-colour}.

\begin{figure}
\begin{centering}
\includegraphics[height=4cm]{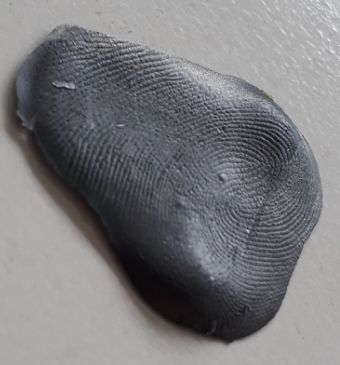}\hspace{0.05\paperwidth}\includegraphics[height=4cm]{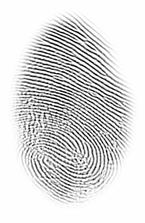}\includegraphics[height=4cm]{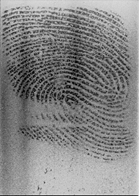}\hspace{0.05\paperwidth}\includegraphics[height=4cm]{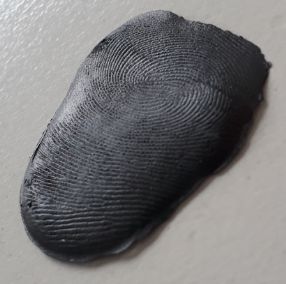}
\par\end{centering}
\caption{Window colour casts with graphite powder in candle wax mould\label{fig:Casts-window-colour-graphite}}
\end{figure}

The second attempt (\textit{Attempt \#8}) was to additionally apply
graphite powder to the mould prior to pouring the window colour into
the mould. Examples of this cast are depicted in Fig. \ref{fig:Casts-window-colour-graphite}.
This approach significantly improved the results over the first one
and the cast worked with all of the tested fingerprint capturing devices.

\paragraph{Fimo (Modelling Clay):}

\begin{figure}
\begin{centering}
\includegraphics[height=4cm]{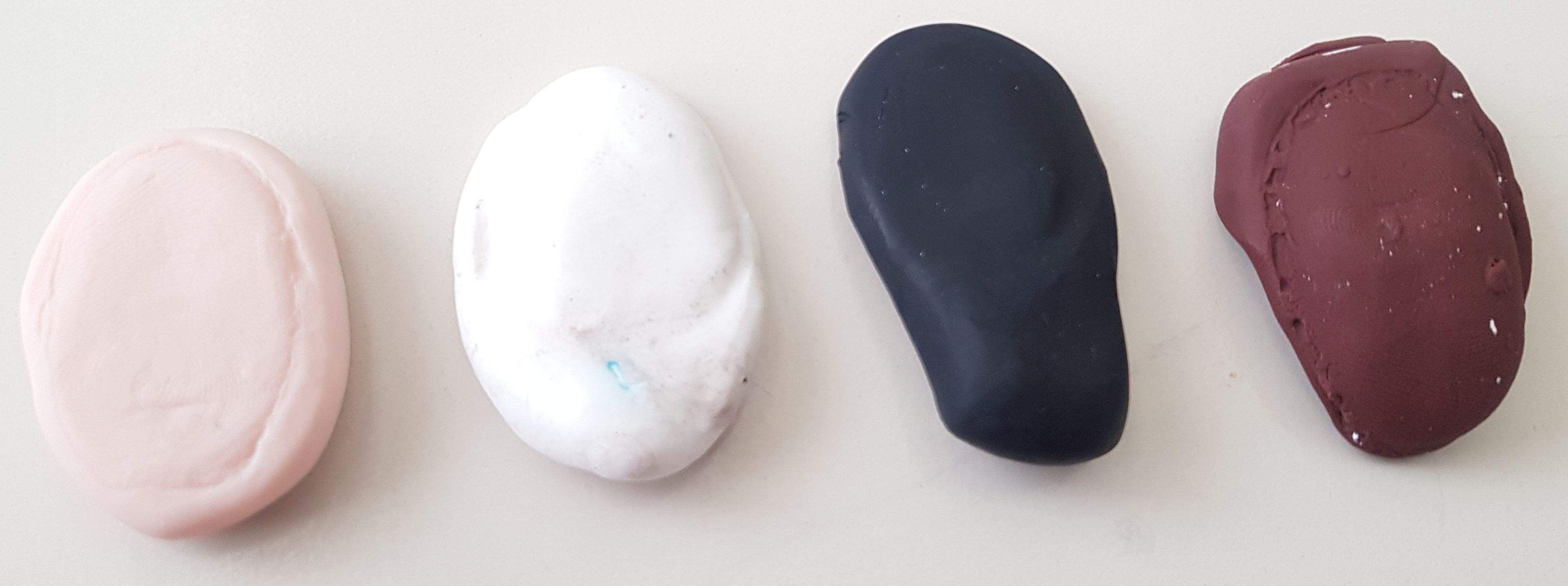}
\par\end{centering}
\caption{Fimo casts\label{fig:Casts-fimo}}
\end{figure}

(\textit{Attempt \#16}) The modelling clay was put into the cast mould
to create a flat and a normal cast. The first attempt was to use the
Fimo cast prior to curing which only worked with the Lumidigm scanner.
The flat ones worked better than the normal, thick ones. The second
attempt was to use the Fimo casts after curing in the over for 30
min at 110\textdegree C and then further curing for 2h at room temperature.
These casts worked worse than the ones prior to curing, again they
only worked with the Lumidign scanner. Several examples of different
Fimo casts can be seen in Fig. \ref{fig:Casts-fimo}.

\paragraph{Silicone:}

\begin{figure}
\begin{centering}
\includegraphics[height=4cm]{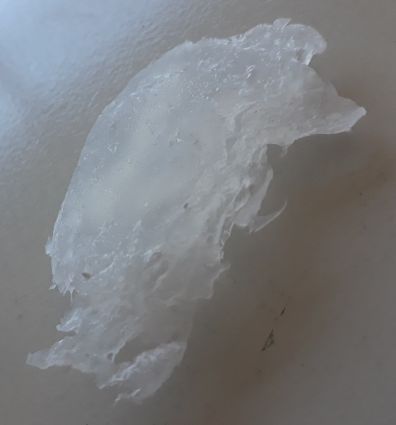}\hspace{0.05\paperwidth}\hspace{0.05\paperwidth}\includegraphics[height=4cm]{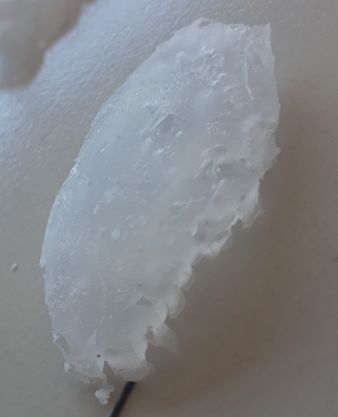}
\par\end{centering}
\caption{Silicone casts\label{fig:Casts-silicone}}
\end{figure}

(\textit{Attempt \#9}) Silicone was poured into the cast mould and
left there 5 days for curing. Fig. \ref{fig:Casts-silicone}shows
some examples of the silicone casts. The problem with this cast is
that it develops many small air bubbles which impact the cast quality.
This cast did work with none of the tested fingerprint capturing devices.

\paragraph{Silicone (Pour):}

\begin{figure}
\begin{centering}
\includegraphics[height=4cm]{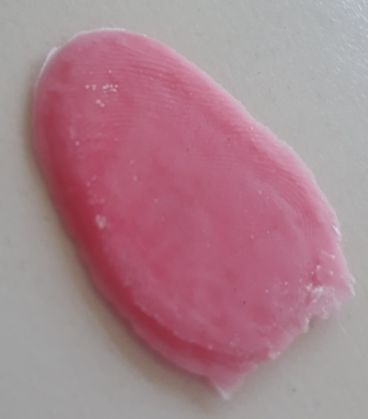}\hspace{0.05\paperwidth}\includegraphics[height=4cm]{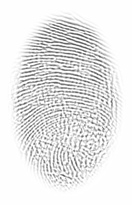}\hspace{0.05\paperwidth}\includegraphics[height=4cm]{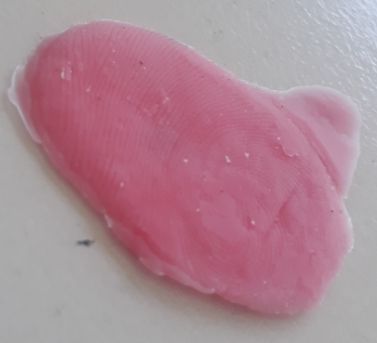}
\par\end{centering}
\caption{Silicone (pour) casts, cast mould\label{fig:Casts-silicone-pour}}
\end{figure}

\begin{figure}
\begin{centering}
\includegraphics[height=4cm]{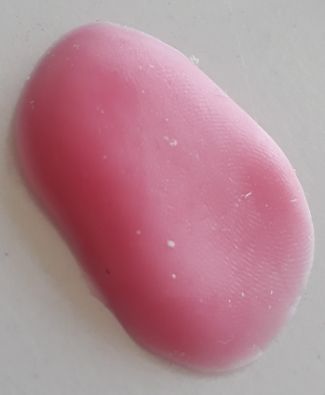}\hspace{0.05\paperwidth}\includegraphics[height=4cm]{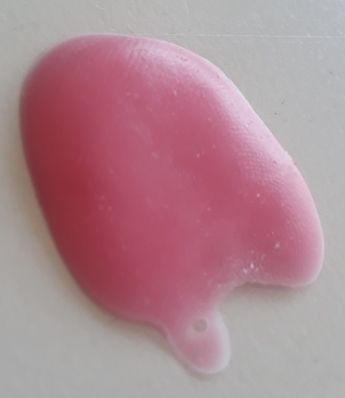}\hspace{0.05\paperwidth}\includegraphics[height=4cm]{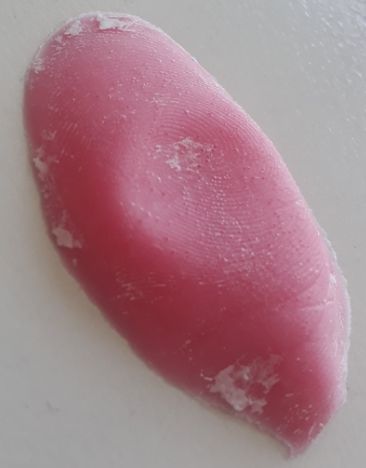}\hspace{0.05\paperwidth}\includegraphics[height=4cm]{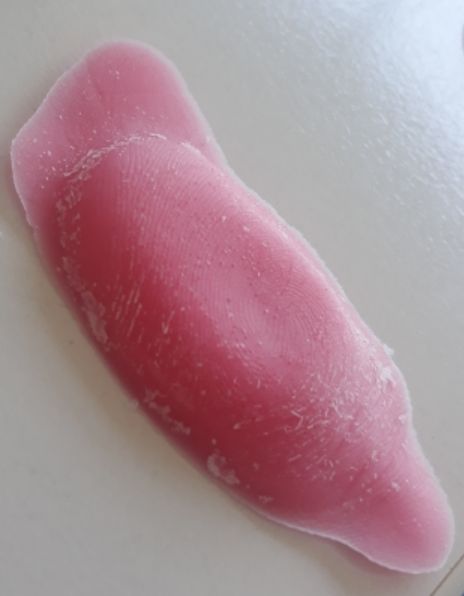}
\par\end{centering}
\caption{Silicone (pour) casts, left: wax mould, right: Play-Doh mould\label{fig:Casts-silicone-pour-wax-playdoh}}
\end{figure}

\begin{figure}
\begin{centering}
\includegraphics[height=4cm]{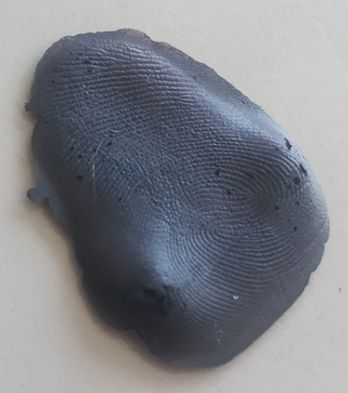}\hspace{0.05\paperwidth}\includegraphics[height=4cm]{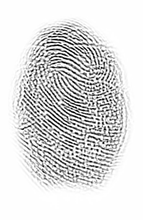}\hspace{0.05\paperwidth}\includegraphics[height=4cm]{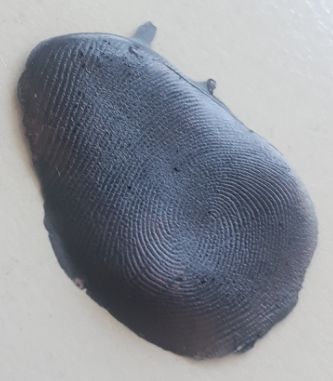}
\par\end{centering}
\caption{Silicone (pour) casts with graphite powder\label{fig:Casts-silicone-pour-graphite}}
\end{figure}

(\textit{Attempt \#17}) This is a different type of silicone but the
casts were done in the same way as the silicone ones using the cast
mould, the wax mould and the Play-Doh mould. The silicone (pour) casts
are depicted in Fig. \ref{fig:Casts-silicone-pour} and Fig. \ref{fig:Casts-silicone-pour-wax-playdoh}.
They worked with the Lumidigm and the thermal fingerprint scanner.
After applying a small layer of graphite, it worked with the passive
capacitive scanner as well. Example images of the graphite powder
layer applied casts can be seen in Fig. \ref{fig:Casts-silicone-pour-graphite}.

\paragraph{Siligum:}

\begin{figure}
\begin{centering}
\includegraphics[height=4cm]{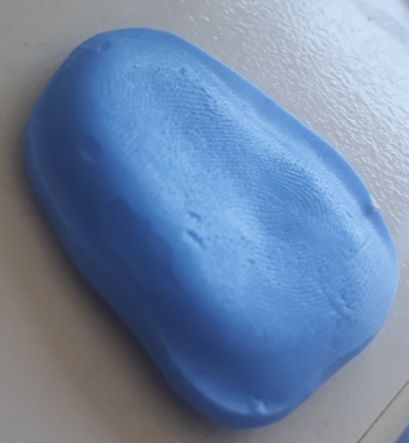}\hspace{0.05\paperwidth}\includegraphics[height=4cm]{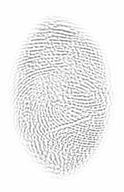}\hspace{0.05\paperwidth}\includegraphics[height=4cm]{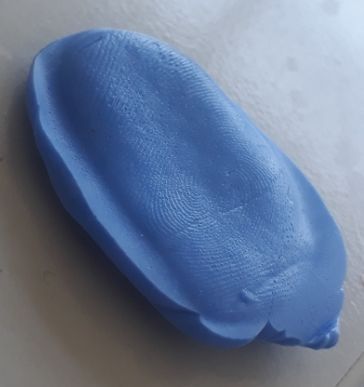}
\par\end{centering}
\caption{Siligum casts\label{fig:Casts-siligum}}
\end{figure}

(\textit{Attempt \#18 and \#25}) Siligum is a dental modelling material.
It was pressed into the cast mould and the left there for about half
an hour to cure. Removal of the cast from the mould was easy and without
any residuals of the mould on the cast. Some example images are shown
in Fig. \ref{fig:Casts-siligum}. This type of cast worked with the
Lumidigm scanner, but none of the other tested devices showed any
reaction.

\paragraph{Formable Art Eraser:}

\begin{figure}
\begin{centering}
\includegraphics[height=4cm]{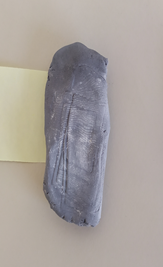}\hspace{0.05\paperwidth}\includegraphics[height=4cm]{Images/Spoofs/Kerzenwachs_1}\hspace{0.05\paperwidth}\includegraphics[height=4cm]{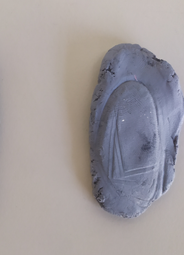}
\par\end{centering}
\caption{Formable art eraser casts\label{fig:Casts-formable-eraser}}
\end{figure}

(\textit{Attempt \#19}) A special kind of eraser that can be formed.
Again it was used in combination with the cast mould by pressing a
piece of the formable art eraser into the mould and removing the cast
from the mould after a few minutes. Fig. \ref{fig:Casts-formable-eraser}
shows some example images. Again this type of cast only worked with
the Lumidigm scanner, while the active capacitive and the thermal
ones showed no reaction and the passive capacitive one detected it
as a spoofed fingerprint. Applying an additional, small layer of graphite
powder did not improve the results.

\paragraph{``Uhu'' Glue:}

\begin{figure}
\begin{centering}
\includegraphics[height=4cm]{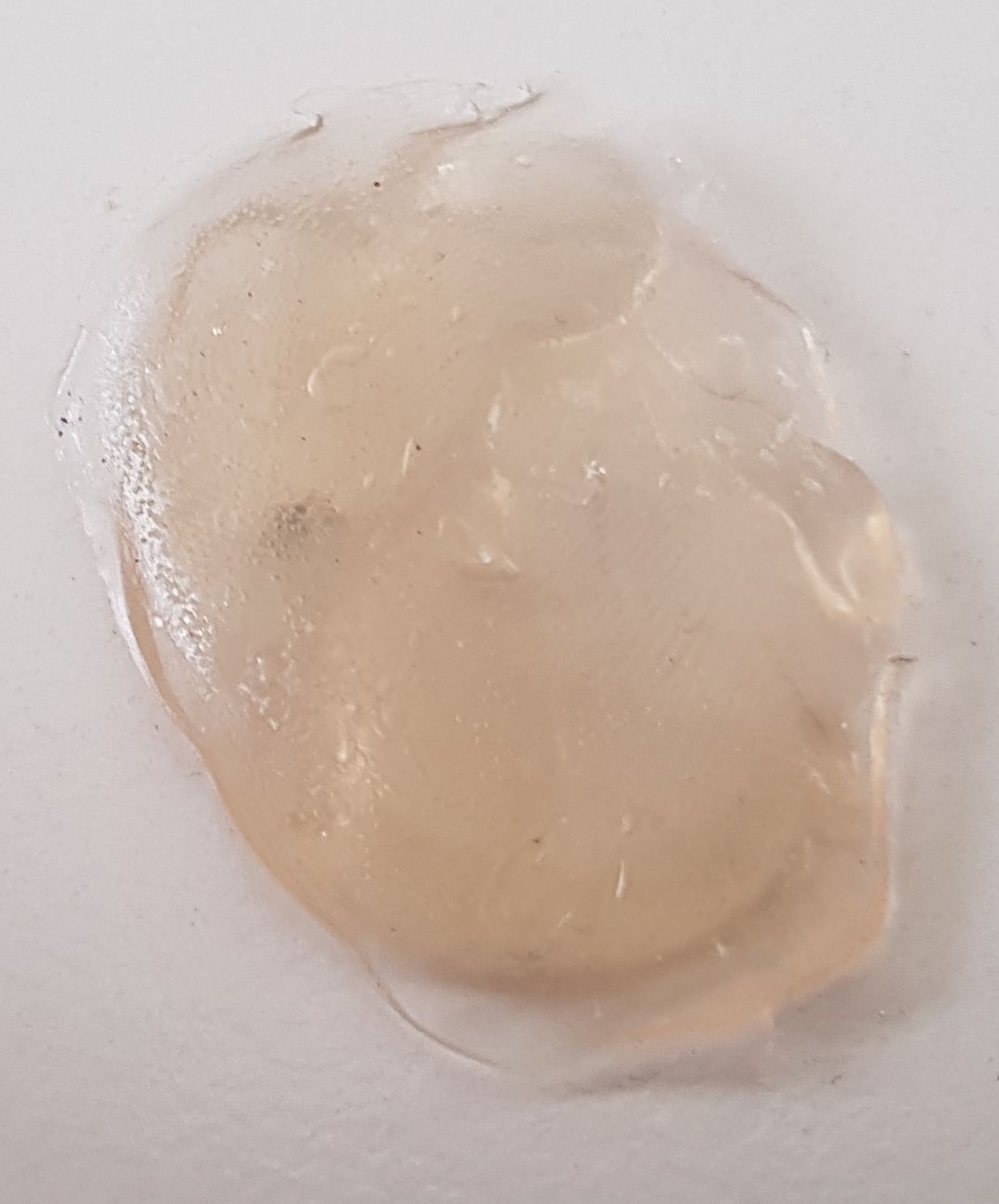}\hspace{0.05\paperwidth}\includegraphics[height=4cm]{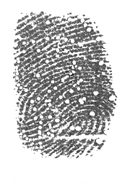}\includegraphics[height=4cm]{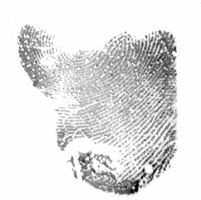}\hspace{0.05\paperwidth}\includegraphics[height=4cm]{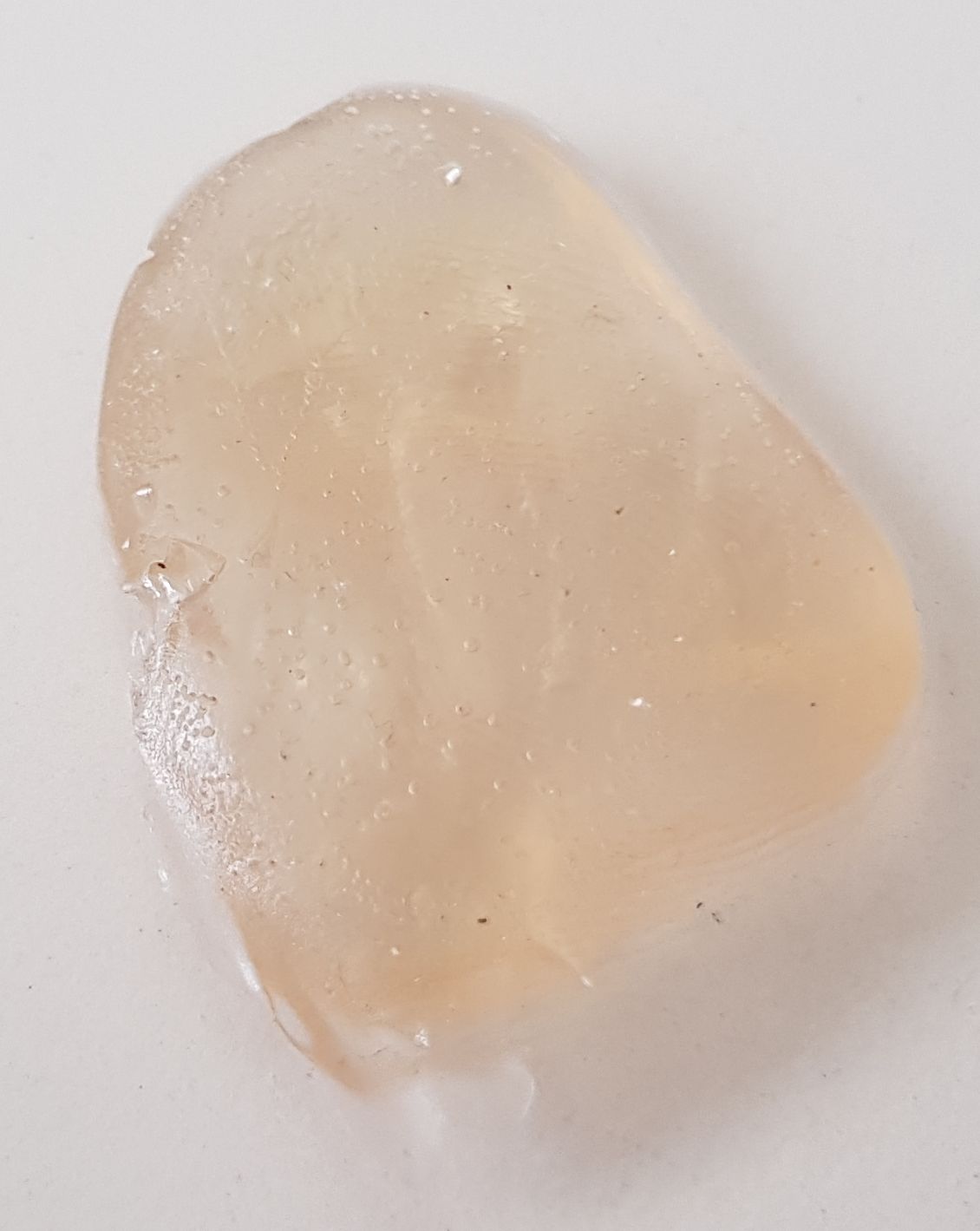}
\par\end{centering}
\caption{Uhu glue casts\label{fig:Casts-Uhu}}
\end{figure}

(\textit{Attempt \#20}) For this attempt, some glue from the brand
``Uhu'' was poured into the cast mould and let there over night
to cure. Otherwise it was removed from the mould to obtain the cast.
Again, a problem with this time of cast are the many air bubbles enclosed
in the cast, lowering the fingerprint sample quality. The cast worked
with the two optical fingerprint scanners as well as with the thermal
one. Fig. \ref{fig:Casts-Uhu} depicts an example of this type of
cast as well as the captured fingerprint samples.

\paragraph{Wood Glue:}

\begin{figure}
\begin{centering}
\includegraphics[height=4cm]{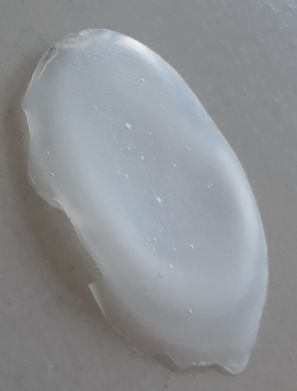}\hspace{0.05\paperwidth}\includegraphics[height=4cm]{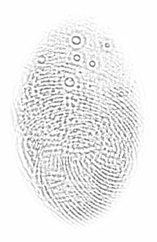}\hspace{0.05\paperwidth}\includegraphics[height=4cm]{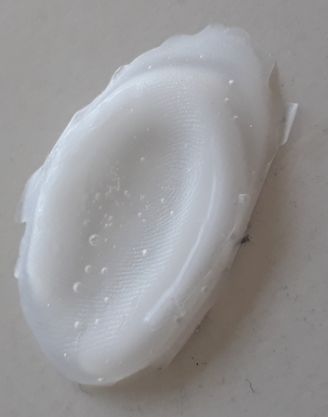}
\par\end{centering}
\caption{Wood glue (thin) casts\label{fig:Casts-woodglue-thin}}
\end{figure}

(\textit{Attempt \#10}) The first attempt was to pour a small amount
of wood glue into the candle wax mould in order to create a thin (3-4
mm) cast. The wood glue was left in the mould for 24h to cure. A second
attempt was to use thicker layer (about 20 mm) which needed a few
days to cure. The thick cast did not work at all, the thin one worked
with the Lumidigm scanner. Fig. \ref{fig:Casts-woodglue-thin} shows
some examples of the thin one. One problem with this type of cast
are the small air bubbles enclosed in the cast.

\begin{figure}
\begin{centering}
\includegraphics[height=4cm]{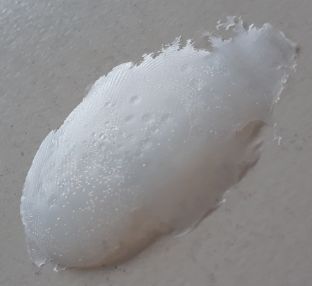}\hspace{0.05\paperwidth}\includegraphics[height=4cm]{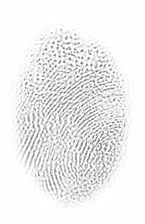}\hspace{0.05\paperwidth}\includegraphics[height=4cm]{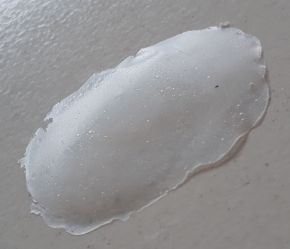}
\par\end{centering}
\caption{Wood glue (extra thin) casts\label{fig:Casts-woodglue-extrathin}}
\end{figure}

\begin{figure}
\begin{centering}
\includegraphics[height=4cm]{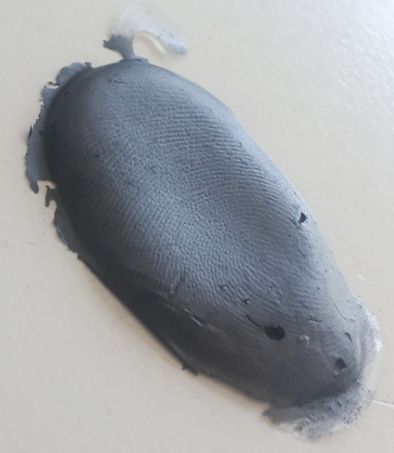}\hspace{0.05\paperwidth}\includegraphics[height=4cm]{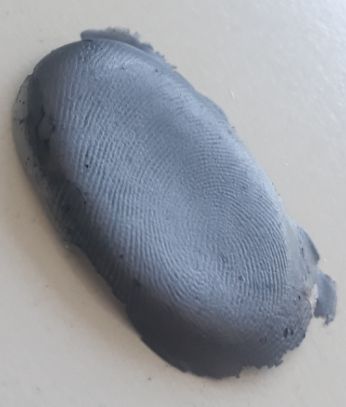}
\par\end{centering}
\caption{Wood glue (extra thin) casts with graphite powder\label{fig:Casts-woodglue-extrathin-graphite}}
\end{figure}

The third attempt (\textit{Attempt \#11}) was to use an even thinner
layer of wood glue (less than 1 mm), again in combination with the
candle wax mould. Some example images of these casts are depicted
in Fig. \ref{fig:Casts-woodglue-extrathin}. For those casts to work
it is necessary to put them onto the finger, otherwise the fingerprint
scanners do not detect the cast. The extra thin version worked with
the Lumidigm and the thermal scanner. By applying a thin layer of
graphite powder (as depicted in Fig. \ref{fig:Casts-woodglue-extrathin-graphite}),
it worked with the active capacitive and the passive capacitive scanner
as well.

\paragraph{Acrylic:}

\begin{figure}
\begin{centering}
\includegraphics[height=4cm]{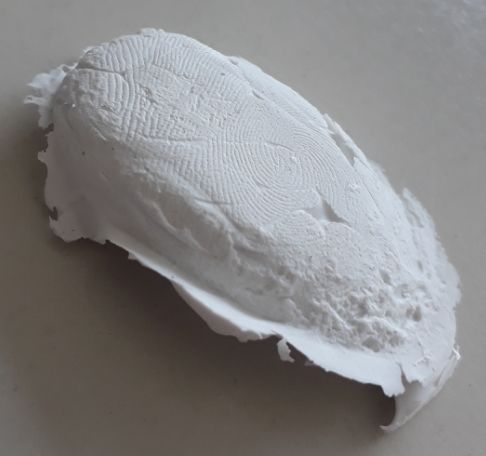}\hspace{0.05\paperwidth}\includegraphics[height=4cm]{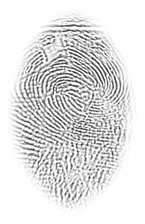}\includegraphics[height=4cm]{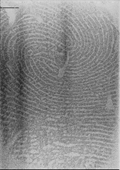}\hspace{0.05\paperwidth}\includegraphics[height=4cm]{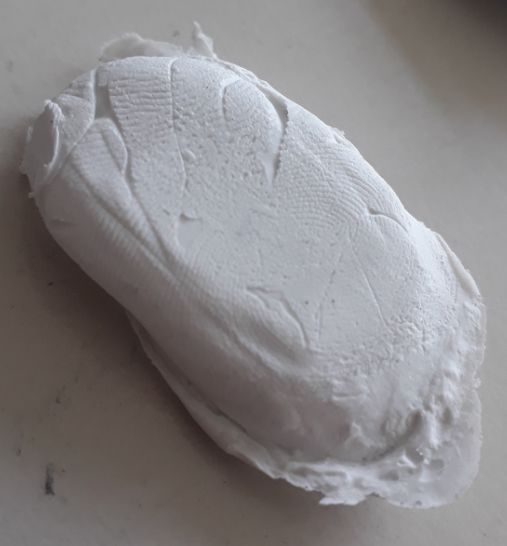}
\par\end{centering}
\caption{Acrylic casts\label{fig:Casts-acrylic}}
\end{figure}

(\textit{Attempt \#12}) The acrylic was put into the cast mould and
left there 24h for curing. Fig. \ref{fig:Casts-acrylic} shows some
example images. As it can be seen, the structures on the cast are
rather coarse and there are some holes, making the surface non homogeneous.
These casts worked with the Lumidigm scanner and the RealScan detected
them as spoofed fingerprints. All other scanners showed no reaction
at all.

\paragraph{Sealing Wax:}

\begin{figure}
\begin{centering}
\includegraphics[height=4cm]{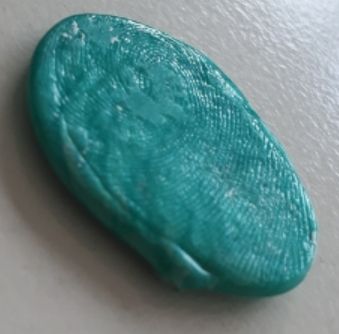}\hspace{0.05\paperwidth}\includegraphics[height=4cm]{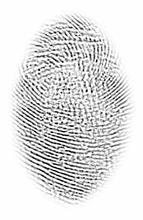}\hspace{0.05\paperwidth}\includegraphics[height=4cm]{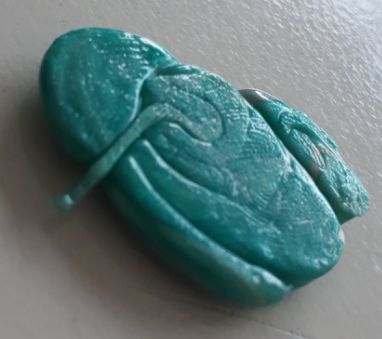}
\par\end{centering}
\caption{Sealing wax casts\label{fig:Casts-sealing-wax}}
\end{figure}

(\textit{Attempt \#13}) The sealing wax was heated in the same way
as the candle wax and poured into the cast mould while it was still
liquid. After curing it was removed from the mould. Examples of this
cast can be seen in Fig. \ref{fig:Casts-sealing-wax}. They again
only worked with the Lumidigm scanner, the RealScan scanner showed
that the ``finger is too dirty'', all other devices showed no reaction
at all.

\paragraph{Construction Adhesive:}

\begin{figure}
\begin{centering}
\includegraphics[height=4cm]{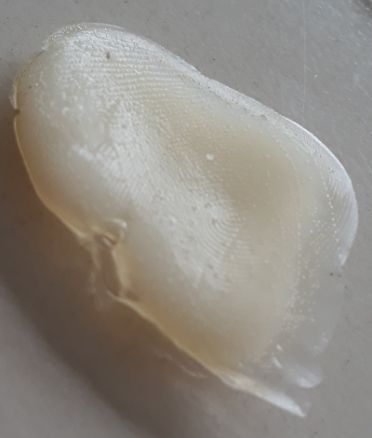}\hspace{0.05\paperwidth}\includegraphics[height=4cm]{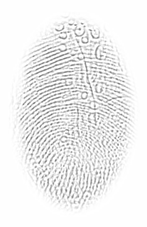}\hspace{0.05\paperwidth}\includegraphics[height=4cm]{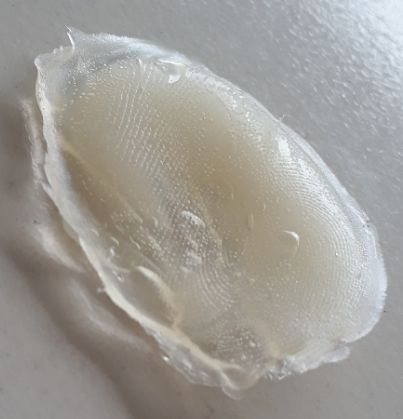}
\par\end{centering}
\caption{Construction adhesive casts\label{fig:Casts-construction-adhesive}}
\end{figure}

(\textit{Attempt \#14}) The construction adhesive was poured into
the candle wax mould and left there about an hour for curing. Again,
a problem with this type of mould are the small air bubbles enclosed
in it, as it can be seen in Fig. \ref{fig:Casts-construction-adhesive}.
These casts worked with the Lumidigm scanner. However, the transparent
areas of the cast were not detected by the scanner. This was improved
by putting a finger behind the cast. The RealScan device detected
it as a spoofed finger, the passive capacitive scanner as bad object,
the active capacitive one showed no reaction and with the thermal
one it worked as well.

\begin{figure}
\begin{centering}
\includegraphics[height=4cm]{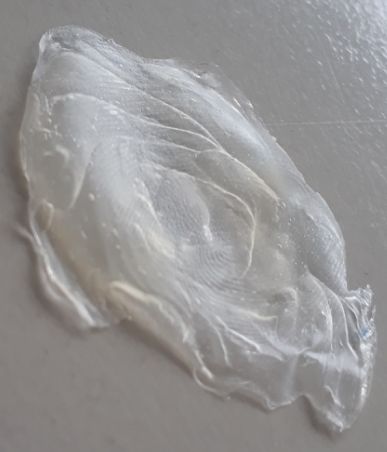}\hspace{0.05\paperwidth}\includegraphics[height=4cm]{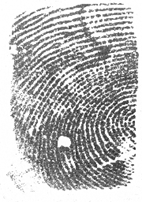}\hspace{0.05\paperwidth}\includegraphics[height=4cm]{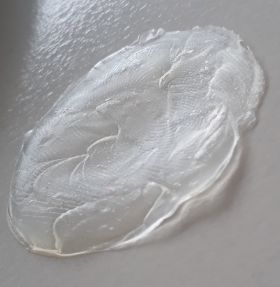}
\par\end{centering}
\caption{Construction adhesive extra thin casts\label{fig:Casts-construction-adhesive-extrathin}}
\end{figure}

The second attempt (\textit{Attempt \#14}) was to use an extra thin
construction adhesive cast as depicted in Fig. \ref{fig:Casts-construction-adhesive-extrathin}.
These casts worked with the Lumidigm, the passive capacitive and the
thermal scanner while the active capacitive one showed no reaction.

\paragraph{Sellotape with Graphite Powder:}

\begin{figure}
\begin{centering}
\includegraphics[height=4cm]{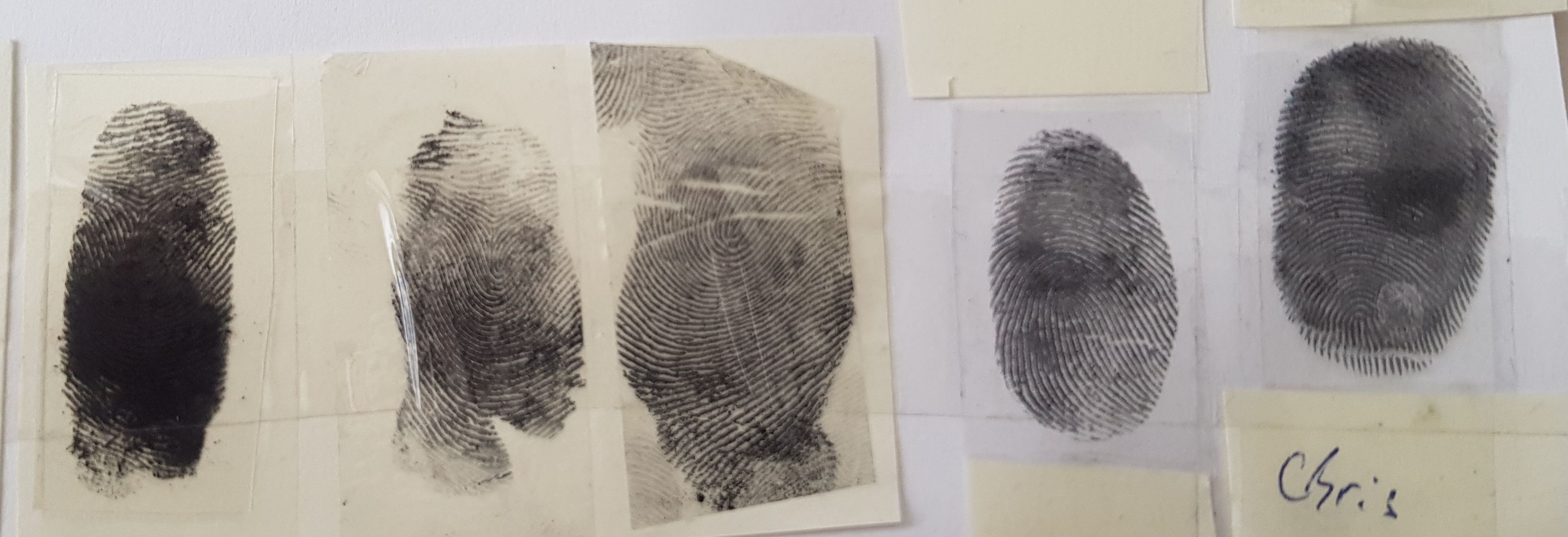}\hspace{0.05\paperwidth}\includegraphics[height=4cm]{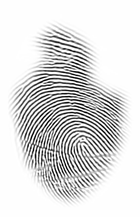}
\par\end{centering}
\caption{Sellotape with graphite powder casts\label{fig:Casts-sellotape-graphite}}
\end{figure}

(\textit{Attempt \#21 and \#26}) For this type of cast, no mould was
necessary. The finger was put on a piece of sellotape which was then
applied with a thin layer of graphite powder. Fig. \ref{fig:Casts-sellotape-graphite}depicts
some example image of this cast. After applying a second layer of
sellotape and covering the back of the tape with the palm of the hand
or some towel, it worked with the Lumidigm scanner but none of the
other fingerprint scanners showed a reaction.

\paragraph{Gelatine with Cornflour and Food Colouring:}

\begin{figure}
\begin{centering}
\includegraphics[height=4cm]{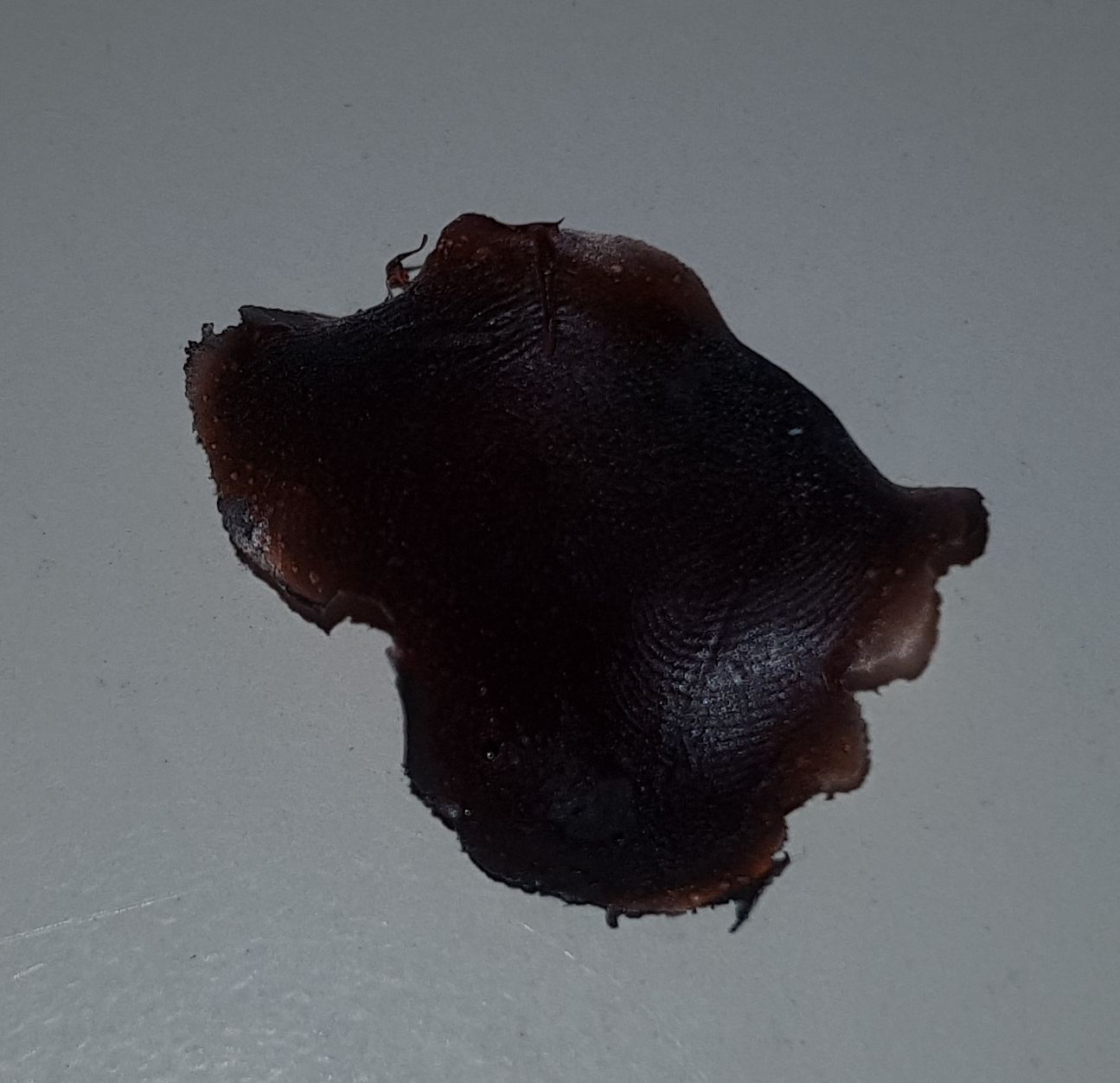}\hspace{0.05\paperwidth}\includegraphics[height=4cm]{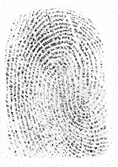}\hspace{0.05\paperwidth}\includegraphics[height=4cm]{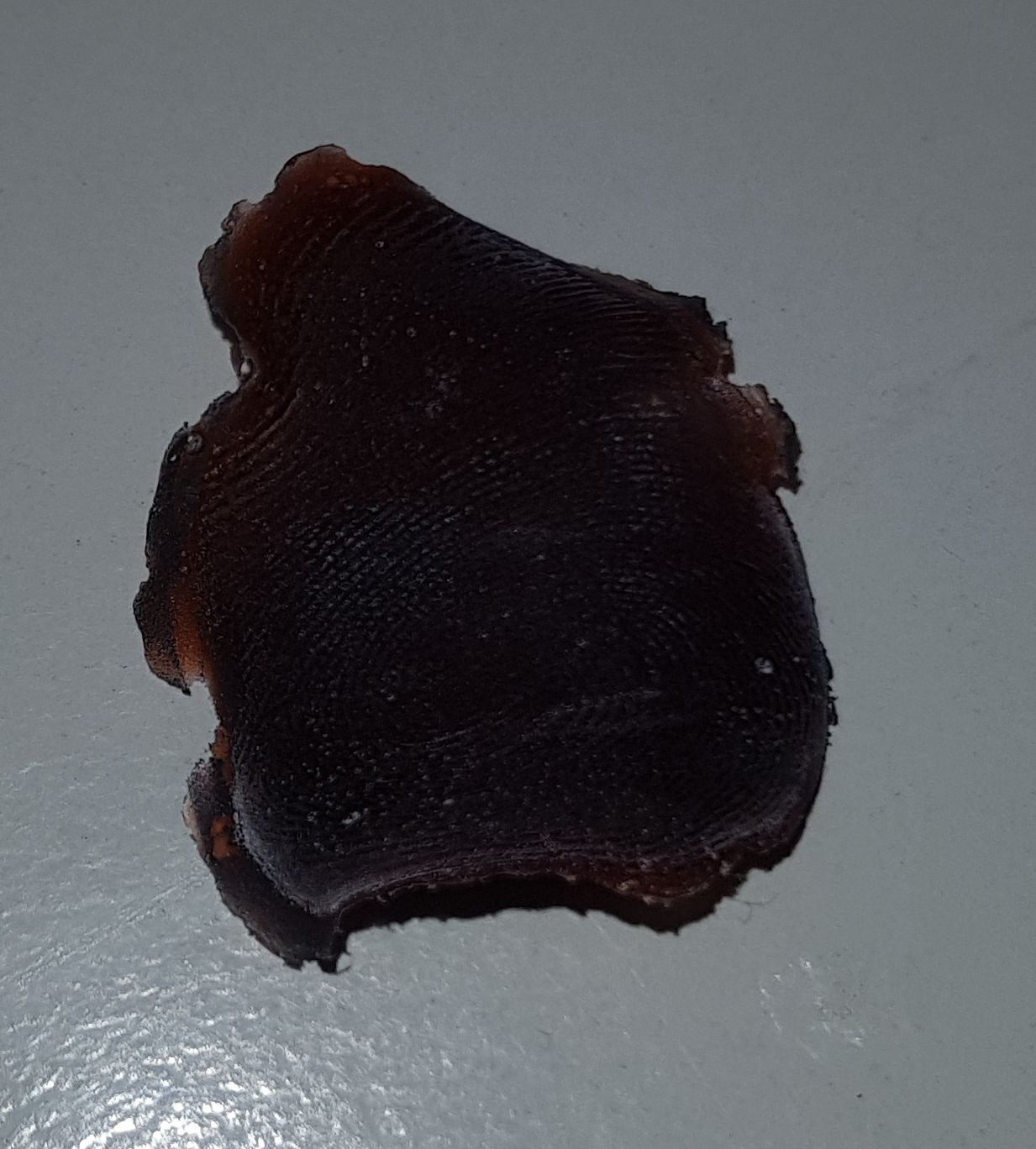}
\par\end{centering}
\caption{Gelatine with cornflour and food colouring casts\label{fig:Casts-gelatine-cornflour-foodcolouring}}
\end{figure}

(\textit{Attempt \#22}) A thin plate of gelatine was applied with
some cornflour and food colouring. Then the finger was placed on the
plate and the imprint was created. Again, no mould is needed. This
cast worked with all fingerprint scanners except the Lumidigm one.
Fig. \ref{fig:Casts-gelatine-cornflour-foodcolouring} shows some
example images.

\paragraph{``Blu Tack'':}

\begin{figure}
\begin{centering}
\includegraphics[height=4cm]{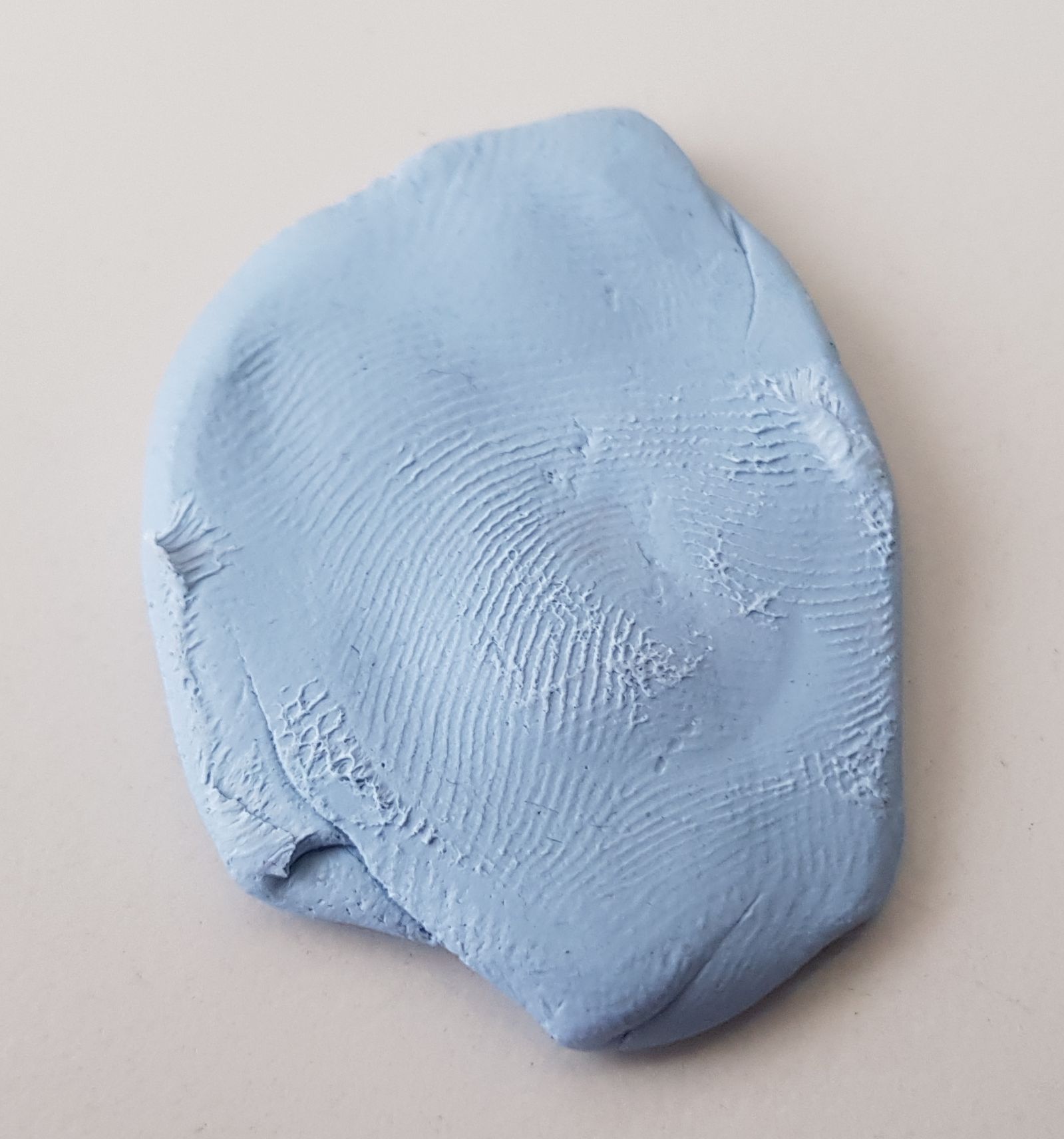}
\par\end{centering}
\caption{Natural latex casts\label{fig:Casts-blutack}}
\end{figure}

(\textit{Attempt \#23}) Blu Tack is kind of an adhesive tape by Scotch.
Again, the finger was just pressed into the piece of Blu Tack, no
mould is needed. An example of this cast can be seen in Fig. \ref{fig:Casts-blutack}.
This type of cast only worked for the Lumidigm scanner again, while
all others showed no reaction at all.

\paragraph{Modelling Clay ``Eberhard'':}

\begin{figure}
\begin{centering}
\includegraphics[height=4cm]{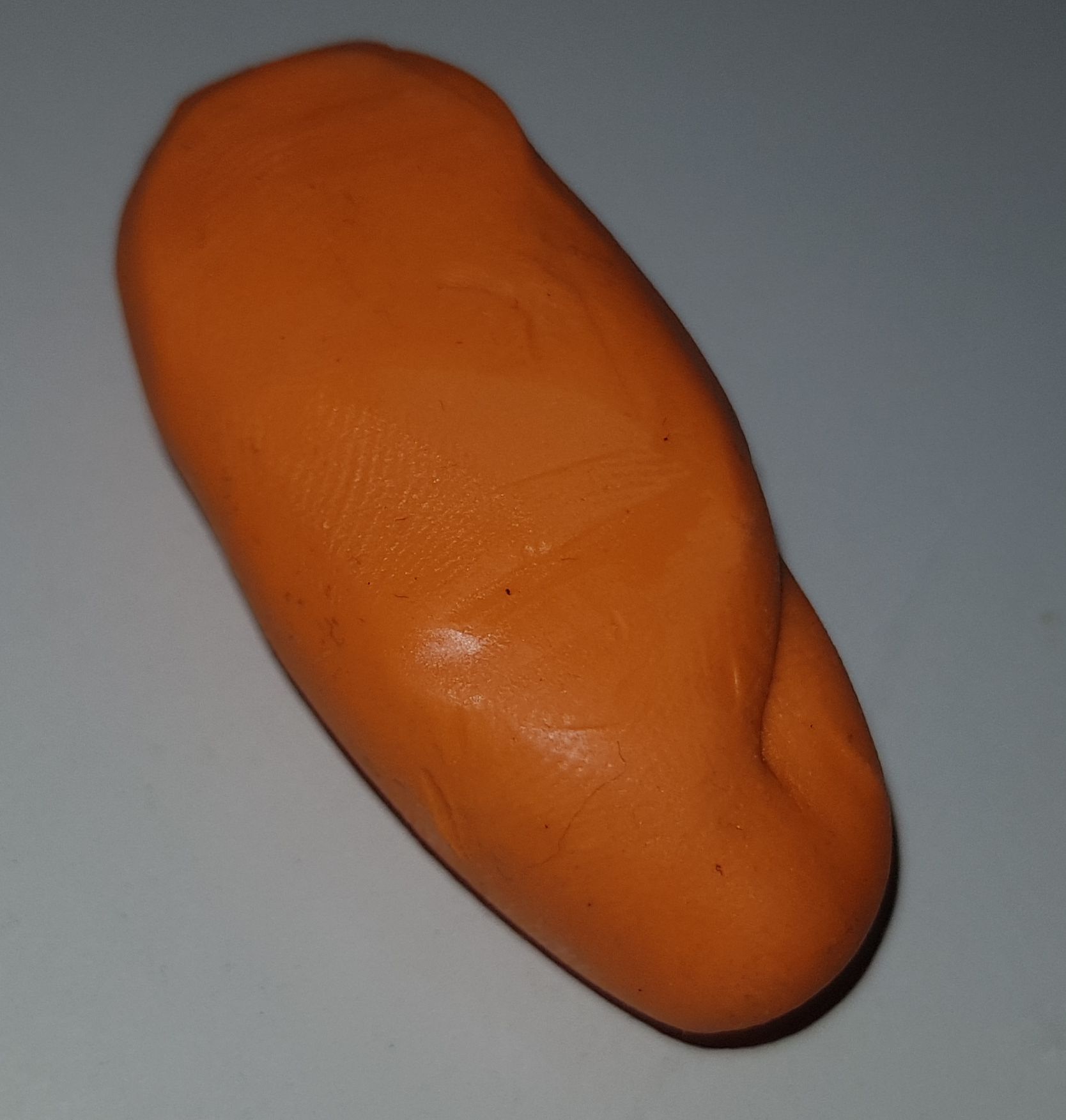}
\par\end{centering}
\caption{Natural latex casts\label{fig:Casts-eberhard}}
\end{figure}

(\textit{Attempt \#24}) The Eberhard modelling clay was pressed into
the cast mould to create the cast. Afterwards it was directly applied
to the different fingerprint scanners. An example of this cast is
depicted in Fig. \ref{fig:Casts-eberhard}. It worked with the two
optical scanners (Lumidigm and RealScan) while all others did not
show any reaction.

\paragraph{Sellotape with Wax and Graphite Powder:}

\begin{figure}
\begin{centering}
\includegraphics[height=4cm]{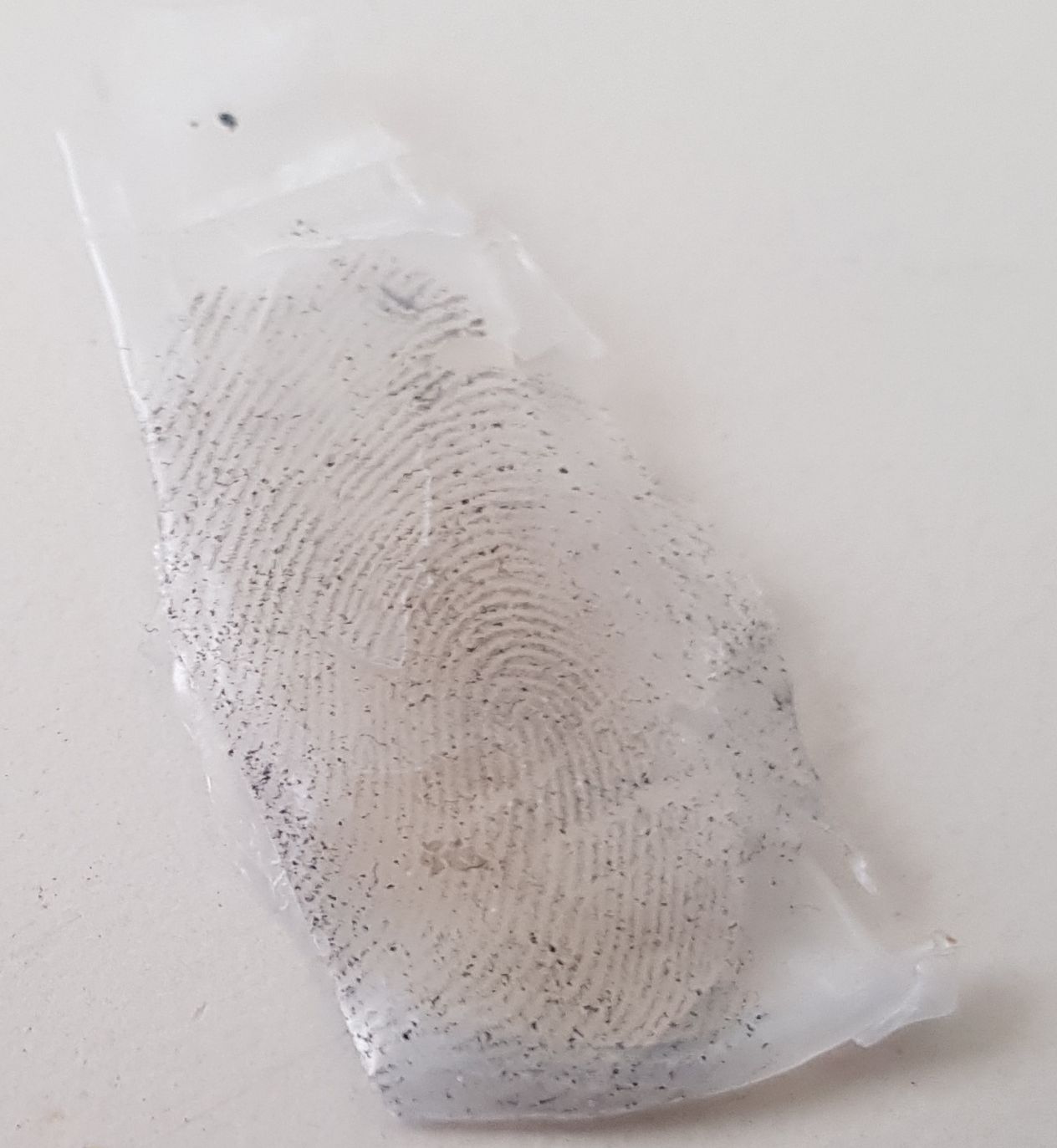}
\par\end{centering}
\caption{Sellotape casts with wax and graphite powder\label{fig:Casts-sellotape-wax-grapihite}}
\end{figure}

(\textit{Attempt \#26}) The finger was again put on a piece of sellotape
which was then applied with a thin layer of graphite powder. Instead
of applying a second layer of tape, a small layer of candle wax was
applied on the backside. Fig. \ref{fig:Casts-sellotape-wax-grapihite}
shows an example of this variant of cast. This variant of the sellotape
cast worked with the Lumidigm scanner, while all other scanners showed
no reaction again.

\paragraph{Uhu Glue with Green Food Colouring}

\begin{figure}
\begin{centering}
\includegraphics[height=4cm]{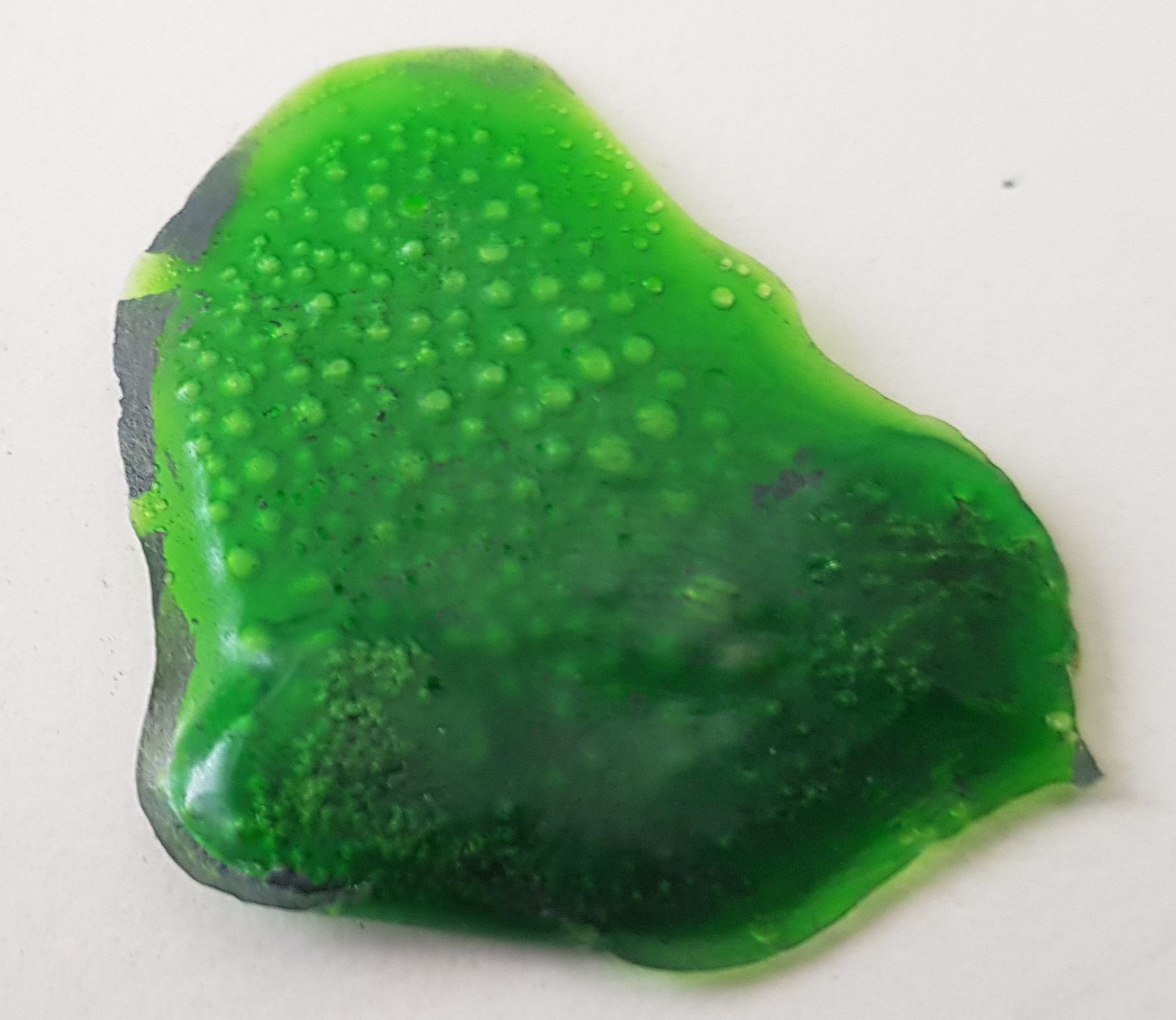}\hspace{0.05\paperwidth}\includegraphics[height=4cm]{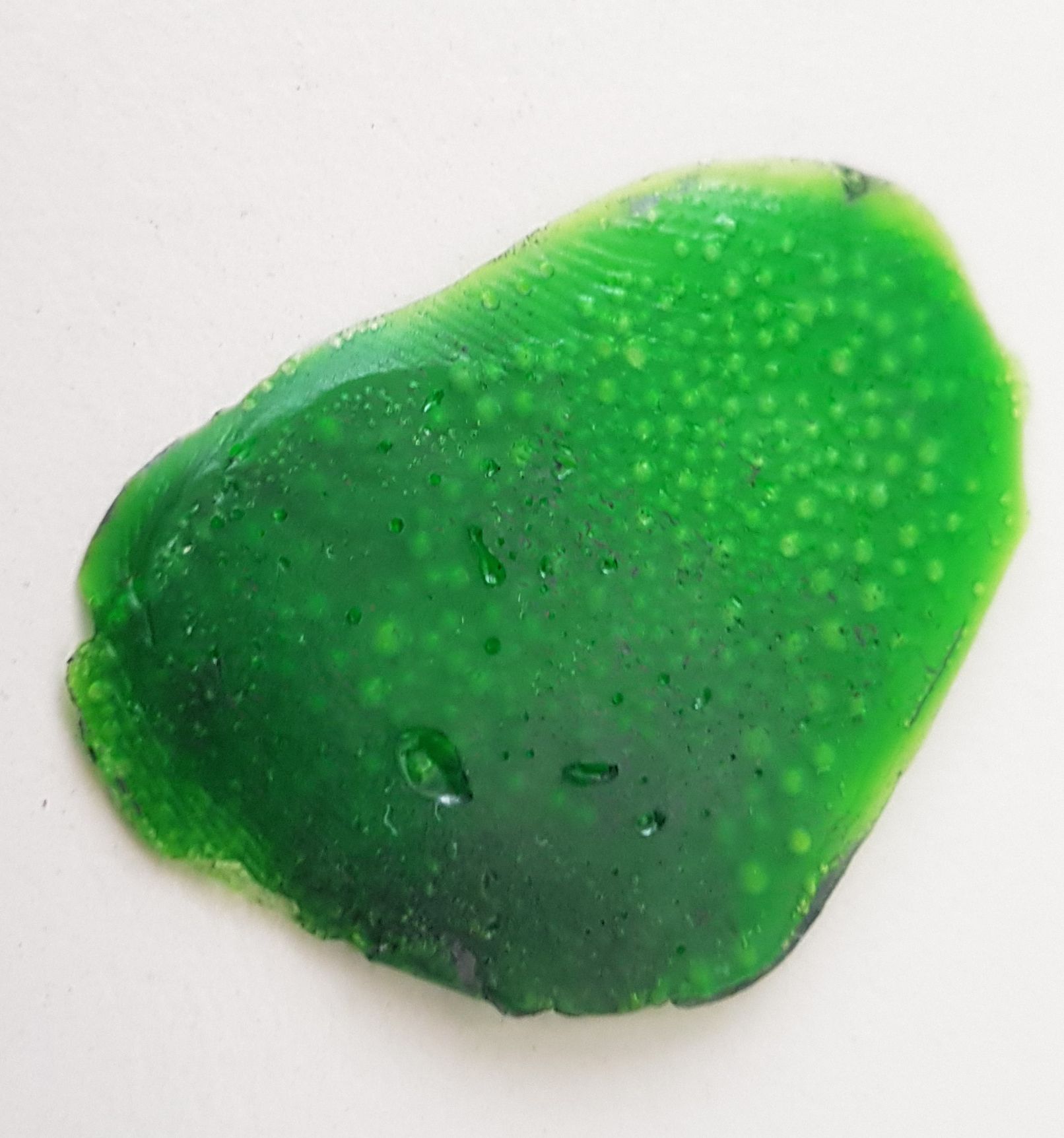}
\par\end{centering}
\caption{Uhu glue casts with green food colouring \label{fig:Casts-Uhu-FoodColouring}}
\end{figure}

(\textit{Attempt \#2}8) The Uhu glue was mixed with green food colouring
and then poured into the Fimo mould and left there a few hours for
curing. Afterwards the casts were removed and presented to the different
fingerprint scanners. The problem with this kind of cast is again
that there are many air bubbles enclosed in the cast. None of them
showed any reaction. Some example images of these casts are shown
in Fig. \ref{fig:Casts-Uhu-FoodColouring}.

\paragraph{Natural Latex:}

\begin{figure}
\begin{centering}
\includegraphics[height=4cm]{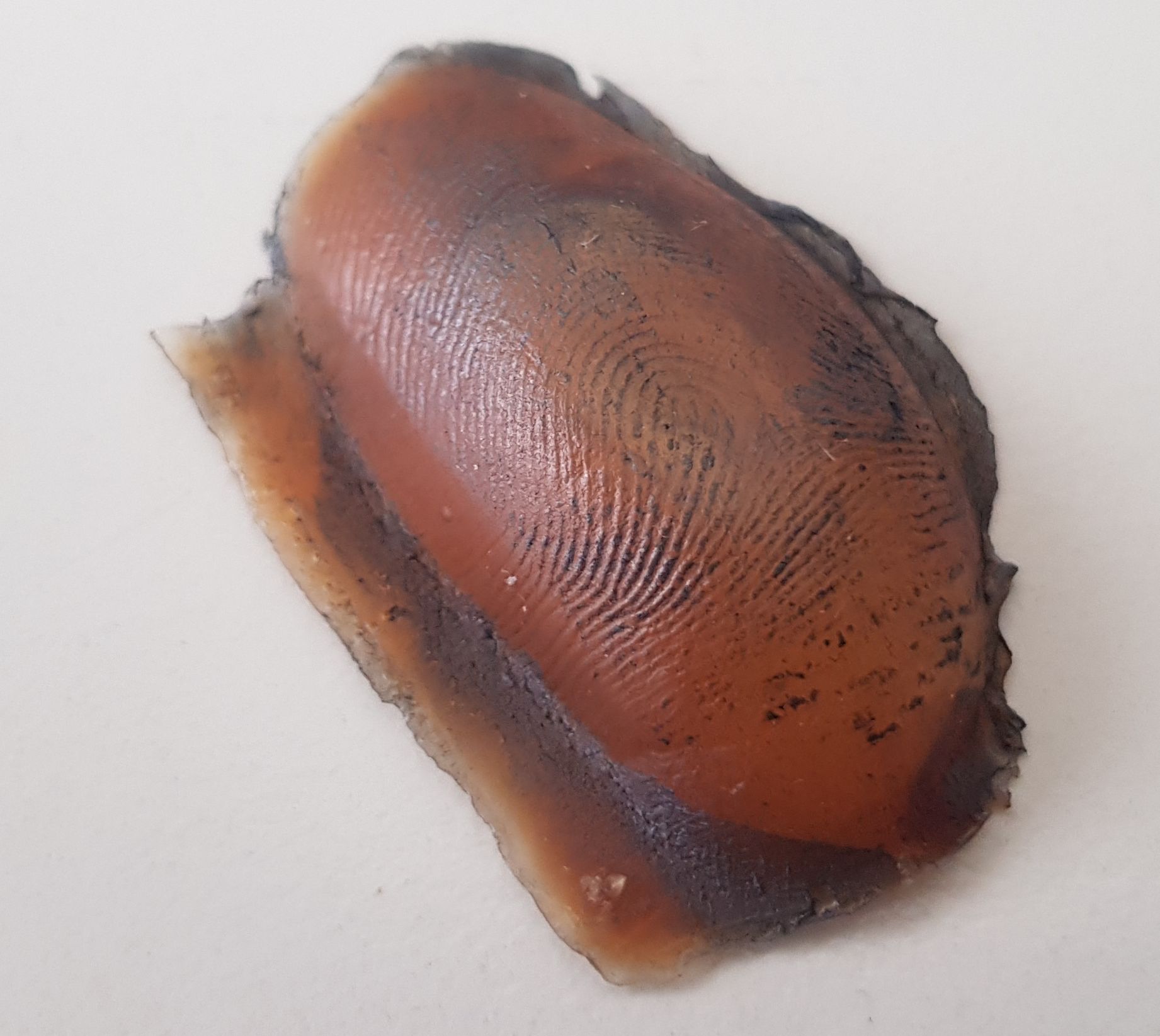}\hspace{0.05\paperwidth}\includegraphics[height=4cm]{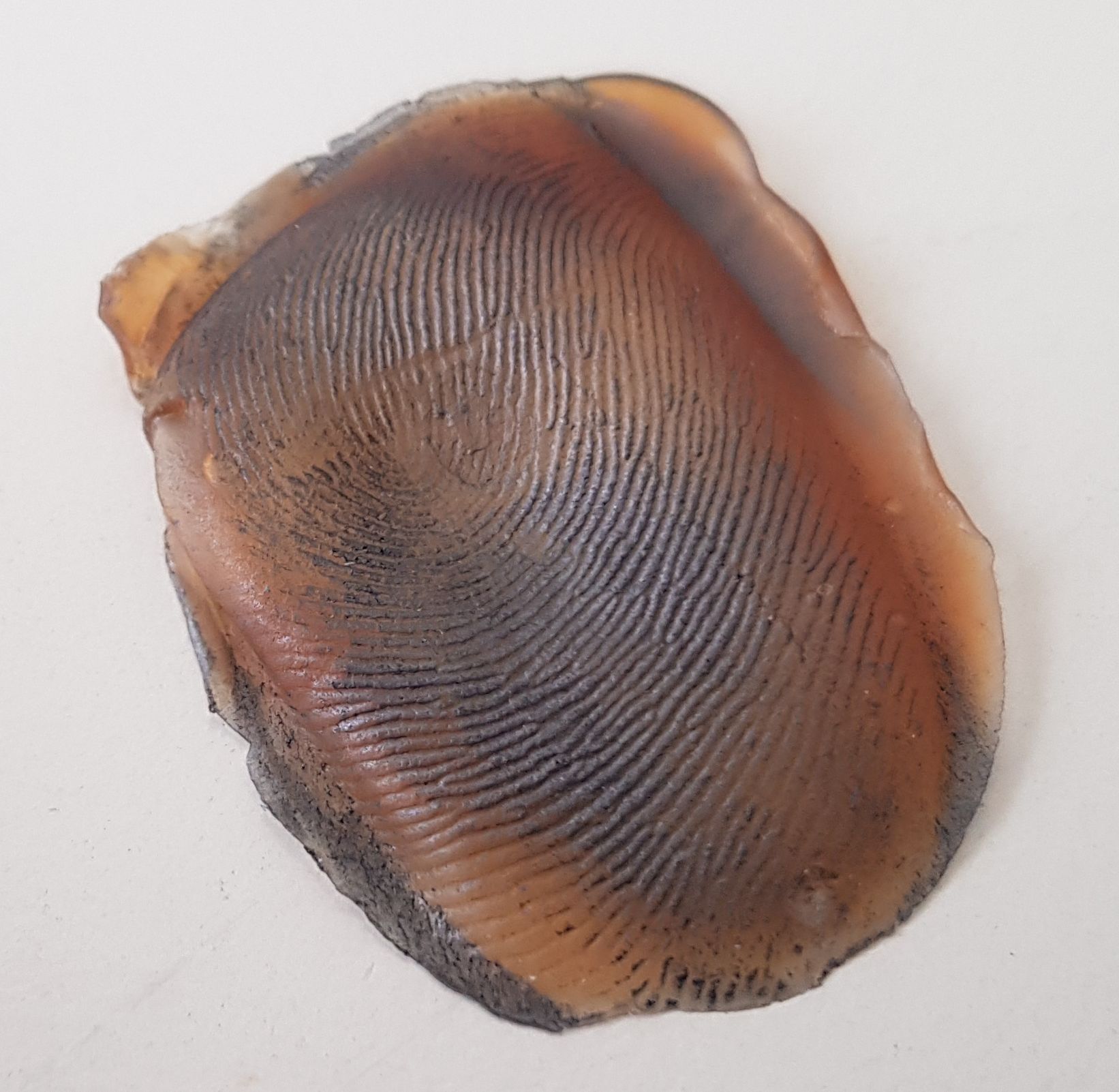}
\par\end{centering}
\caption{Natural latex casts\label{fig:Casts-natural-latex}}
\end{figure}

(\textit{Attempt \#30}) The natural latex was poured into the cast
mould and left there for about 24h to cure. Afterwards it was removed
from the mould. Fig. \ref{fig:Casts-natural-latex} shows some example
images. These casts worked for all fingerprint scanners except the
thermal one.

\subsection{Spoofed Fingerprint Quality\label{subsec:Spoofed-Fingerprint-Quality}}

Table \ref{tab:Fingerprint-quality-results} lists the fingerprint
quality values for the different presentation attack (spoofing artefact)
attempts and the five tested fingerprint scanners. The quality is
evaluated using Verifinger's built in fingerprint quality measure,
which quantifies the quality as integer values from 0 to 100, with
0 indicating the lowest possible quality and 100 indicating the best
possible one. A ``-'' indicates that no template could be extracted,
hence there is no quality value. As the table reveals, most spoofing
attempts worked for the Lumidigm V311 scanner, followed by the NB-3010-U
and the Verifi P5000 and IB Columbo while only two of the attempts
worked with the Realscan G1. Most of the working attempts achieved
rather high quality values (>50) which indicated a good quality of
the artificial fingerprint artefacts. The overall best performing
attempt was \#30, corresponding to the natural latex casts. Attempt
\#8, \#15, \#17 and \#21 also worked well for some of the fingerprint
scanners. Hence, we decided to include only those five attempts in
the matching performance experiments which are described in the next
subsection.

\begin{table}[H]
\centering{}\caption{Fingerprint quality results of the different fingerprint spoofs (different
casts). The quality values are in the range from 0-100, where 0 corresponds
to the lowest quality and 100 to the highest possible quality. The
first column indicates the attempt nr., which is mentioned for each
attempt in subsection \ref{subsec:Molds}. - means that there was
either no reaction at all or the template could not be created due
to low quality or spoof detected. The best working type of cast is
highlighted in \textbf{bold}. \label{tab:Fingerprint-quality-results}}
\begin{tabular}{|c|c|c|c|c|c|}
\hline 
 & \multicolumn{5}{c|}{Fingerprint Quality (range: 0-100)}\tabularnewline
\hline 
Attempt \# & Lumidigm V311 & Realscan G1 & IB Columbo & Verifi P5000 & NB-3010-U\tabularnewline
\hline 
1 & 70 - 85 & - & - & - & -\tabularnewline
\hline 
2 & 60 & - & - & - & -\tabularnewline
\hline 
3 & 80 - 86 & - & - & - & -\tabularnewline
\hline 
4 & 65 & - & - & - & -\tabularnewline
\hline 
5 & 75 & - & - & - & -\tabularnewline
\hline 
6 & 50 & - & - & - & -\tabularnewline
\hline 
7 & - & - & - & - & 40 - 69\tabularnewline
\hline 
8 & 80 - 93 & - & 69 - 74 & 60 - 91 & -\tabularnewline
\hline 
9 & - & - & - & - & -\tabularnewline
\hline 
10 & 42 - 61 & - & - & - & -\tabularnewline
\hline 
11 & 40 - 90 & - & 80 & 70 & 40 - 50\tabularnewline
\hline 
12 & 50 - 65 & - & 60 & 65 & -\tabularnewline
\hline 
13 & 96 & - & - & - & -\tabularnewline
\hline 
14 & 60 - 71 & - & - & - & 44 - 50\tabularnewline
\hline 
15 & 42 - 97 & - & - & 65 & 80 - 85\tabularnewline
\hline 
16 & 60 - 80 & - & - & - & -\tabularnewline
\hline 
17 & 80 - 92 & - & - & 54 & 50\tabularnewline
\hline 
18 & 56 - 62 & - & - & - & -\tabularnewline
\hline 
19 & 45 - 59 & - & - & - & -\tabularnewline
\hline 
20 & 72 & 73 & - & - & 65 -75\tabularnewline
\hline 
21 & 85 - 99 & - & - & 73 - 81 & -\tabularnewline
\hline 
22 & - & - & 60 & 49 - 59 & 66 - 73\tabularnewline
\hline 
23 & 74 & - & - & - & -\tabularnewline
\hline 
24 & 86 & 79 & - & - & -\tabularnewline
\hline 
25 & 74 - 79 & - & - & - & 53\tabularnewline
\hline 
26 & 68 & - & - & 48 & -\tabularnewline
\hline 
27 & 90 & - & - & - & -\tabularnewline
\hline 
28 & - & - & - & - & -\tabularnewline
\hline 
29 & 94 &  &  &  & \tabularnewline
\hline 
\textbf{30} & \textbf{70 - 100} & \textbf{67 - 95} & \textbf{66 - 85} & \textbf{75 - 83} & \textbf{70 - 73}\tabularnewline
\hline 
\end{tabular}
\end{table}

\subsection{Spoofed Fingerprint Matching Performance\label{subsec:Spoofed-Fingerprint-Matching}}

Fake fingerprint samples from about 15 different subjects have been
created using the window colour (Attempt \#8), the extra thin construction
adhesive (Attempt \#15), the silicone (pour) (Attempt \#17), the Sellotape
with graphite powder (Attempt \#21) and the natural latex (Attempt
\#30) as those attempts achieved the highest fingerprint quality values
among the tested fingerprint scanners. Based on those spoofed fingerprint
samples, a matching experiment has been conducted, which is summarised
as matching score values using Neurotechnology Verifinger fingerprint
matcher in Table \ref{tab:Matching-results-spoofs}. The matching
scores range from 0 to about 2000, where 0 means no match and higher
numbers indicate higher similarity. ``-'' means that no match was
found. Typical genuine matches (original sample vs. original sample)
are in the range of 50 - 500. The best performing match scores per
fingerprint scanner are highlighted in \textbf{bold}. The selected
user IDs represent the best and worst user in terms of matching score
per scanner and spoofing attempt. As it can be seen, all of the listed
scores (except for user ID 1, Attempt \#21) achieved match scores
higher than 50, i.e. they would have been accepted as genuine user
in a real system (given that the that match threshold is set at an
usual level).

\begin{table}[H]
\begin{centering}
\caption{Matching results (Neurotechnology Verifinger) for the selected best
performing spoofing attempts. - means that there was no match found.
Best matching scores per fingerprint scanner are highlighted \textbf{bold}.\label{tab:Matching-results-spoofs}}
\par\end{centering}
\centering{}%
\begin{tabular}{|c|c|c|c|c|c|c|}
\hline 
Attempt \# & User ID & Lumidigm V311 & Realscan G1 & IB Columbo & Verifi P5000 & NB-3010-U\tabularnewline
\hline 
8 & 1 & \textbf{202} & - & - & \textbf{186} & 193\tabularnewline
\hline 
8 & 2 & 83 & - & - & \textbf{186} & -\tabularnewline
\hline 
15 & 1 & 110 & - & - & - & 91\tabularnewline
\hline 
17 & 1 & 117 & \textbf{146} & - & - & 152\tabularnewline
\hline 
17 & 2 & 64 & 96 & - & - & 166\tabularnewline
\hline 
21 & 1 & - & - & - & - & -\tabularnewline
\hline 
21 & 2 & 72 & - & - & - & -\tabularnewline
\hline 
30 & 1 & 188 & - & - & 159 & 154\tabularnewline
\hline 
30 & 2 & 121 & - & - & 99 & \textbf{232}\tabularnewline
\hline 
30 & 3 & 126 & - & - & - & -\tabularnewline
\hline 
30 & 3 & 76 & - & - & - & -\tabularnewline
\hline 
\end{tabular}
\end{table}

\subsection{Results Summary and Discussion}

It turned out, that especially the multispectral Lumidigm V311 fingerprint
scanner, which should be more resistant to presentation attacks than
the others, was the easiest one to spoof, followed by the thermal
scanner NB-3010-U from Next Biometrics, but the manufacturer did not
claim any increased spoofing resistance for this scanner. The passive
capacitive sensor Verifi P5000 was harder to spoof than the others,
mainly due to its capacitive principle. The active capacitive one,
the Integrated Biometrics Columbo was even harder than the passive
capacitive one, again due to its capturing principle requiring an
active electrical field between the object on the sensor surface and
the sensor surface itself. Surprisingly, none of the fabricated spoofs
was able to achieve valid match scores if compared against the bona
fide samples. Hence, even though the scanner itself accepted the spoofs,
we were not able to successfully fool the whole system as we would
have been rejected during the matching step. However, active capacitive
scanners are rarely used in practical applications due to their higher
costs compared to the passive capacitive ones. The Suprema RealScan
G1, which is another optical sensor, was hard to spoof as well but
we still able to capture some of the spoofing attempts and even successfully
match them against bona fide samples despite the integrated spoofing
detection built into this device according to the manufacturer.

Regarding the types of spoofing materials, the most promising one
was a cast or Fimo/Cernit mould in combination with natural latex
casts combined with a thin layer of graphite powder (especially to
spoof the capacitive sensors). The even simpler Sellotape approach
in combination with some graphite powder also turned out to work really
well. The other 3 well performing approaches are basically just replacing
the cast material by either silicone (pour), construction adhesive
or window colour while still maintaining the Fimo/Cernit mould for
creating the casts.

\section{Conclusion\label{sec:Conclusion}}

Several different types of fake fingerprint representations have been
produced and evaluated using five common commercial-off-the-shelf
fingerprint scanner devices. The evaluation based on the fingerprint
sample image quality as well as the matching scores between the fake
representations and the real fingerprint samples revealed that all
but one of the tested fingerprint scanners are susceptible to presentation
attacks using the tested materials (successful template creation and
matching). The only fingerprint scanner we were not able to spoof
is the Integrated Biometrics Columbo, using an active capacitive sensing
technique. Regarding the different materials to create the fake representations,
natural latex, silicone (pour) and window colours turned out to work
best in terms of successful template creation (circumventing the presentation
attack detection mechanisms of the scanner devices) and matching.
Hence, it is confirmed once more that presentation attacks using cheap
materials and simple methods to create the fake representations still
pose a severe threat to fingerprint recognition.

\printbibliography
\end{document}